\documentclass{article}

\usepackage{microtype}
\usepackage{graphicx}
\usepackage{subfigure}
\usepackage{booktabs} 
\usepackage{longtable}
\usepackage{rotating} 
\usepackage{multirow}

\usepackage{hyperref}

\usepackage[accepted]{arxiv_icml}

\usepackage{amsmath}
\usepackage{amssymb}
\usepackage{mathtools}
\usepackage{amsthm}
\usepackage{enumitem}

\usepackage[capitalize,noabbrev]{cleveref}

\theoremstyle{plain}
\newtheorem{theorem}{Theorem}[section]

\newtheorem{corollary}[theorem]{Corollary}
\theoremstyle{definition}
\newtheorem{definition}[theorem]{Definition}

\theoremstyle{remark}

\usepackage[textsize=tiny]{todonotes}

\icmltitlerunning{Learning Survival Distributions with the Asymmetric Laplace Distribution}

\begin{document}

\twocolumn[
\icmltitle{Learning Survival Distributions with the Asymmetric Laplace Distribution}



\icmlsetsymbol{equal}{*}

\begin{icmlauthorlist}
\icmlauthor{Deming Sheng}{duke}
\icmlauthor{Ricardo Henao}{duke}
\end{icmlauthorlist}

\icmlaffiliation{duke}{Department of Electrical and Computer Engineering, Duke University}
\icmlcorrespondingauthor{Deming Sheng}{deming.sheng@duke.edu}
\icmlcorrespondingauthor{Ricardo Henao}{ricardo.henao@duke.edu}





\icmlkeywords{Machine Learning, ICML}

\vskip 0.3in
]



\printAffiliationsAndNotice{}  

\begin{abstract}
Probabilistic survival analysis models seek to estimate the distribution of the future occurrence (time) of an event given a set of covariates.
In recent years, these models have preferred nonparametric specifications that avoid directly estimating survival distributions via discretization.
Specifically, they estimate the probability of an individual event at fixed times or the time of an event at fixed probabilities (quantiles), using supervised learning.
Borrowing ideas from the quantile regression literature, we propose a parametric survival analysis method based on the Asymmetric Laplace Distribution (ALD).
This distribution allows for closed-form calculation of popular event summaries such as mean, median, mode, variation, and quantiles.
The model is optimized by maximum likelihood to learn, at the individual level, the parameters (location, scale, and asymmetry) of the ALD distribution.
Extensive results on synthetic and real-world data demonstrate that the proposed method outperforms parametric and nonparametric approaches in terms of accuracy, discrimination and calibration.
\end{abstract}

\section{Introduction}
Survival models\cite{nagpal2021deep}, also known as time-to-event models, are statistical frameworks designed to predict the time until a specific event of interest occurs, given a set of covariates.
These models are particularly valuable in situations where the timing of the event is crucial and often subject to {\em censoring}, which means that in some cases the event has not yet occurred or remains unobserved by the end of the data collection period.
The flexibility and adaptability of survival models have led to their widespread application in various fields, including engineering \cite{lai2006stochastic}, finance \cite{gepp2008role}, marketing \cite{jung2012product}, and, notably, healthcare \cite{zhang2017mining, voronov2018data, lanczky2021web, emmerson2021understanding}.

Survival models can be broadly categorized into parametric, semiparametric, and nonparametric methods, each offering unique strengths depending on the characteristics of the data and the underlying assumptions.
Parametric survival models assume that survival times follow a specific statistical distribution, enabling explicit mathematical modeling of the survival function.
Common examples include the exponential distribution for constant hazards rates \cite{feigl1965estimation}, the Weibull distribution for flexible hazards rate modeling \cite{scholz1996maximum}, and the log-normal distribution for positively skewed survival times \cite{royston2001lognormal}.
Semiparametric methods, such as the Cox proportional hazards model \cite{cox1972regression}, assume a proportional hazards structure without specifying a baseline hazard distribution, which offers robustness and interpretability.
Nonparametric methods, including the Kaplan-Meier estimator \cite{kaplan1958nonparametric} and the Nelson-Aalen estimator \cite{aalen1978nonparametric}, rely solely on observed data, avoiding distributional assumptions while directly estimating survival and hazards (risk) functions.

More recently, neural networks have significantly advanced survival models across parametric, semiparametric, and nonparametric settings.
In parametric methods, LogNorm MLE \cite{hoseini2017comparison} enhances parameter estimation for log-normal distributions.
Semiparametric approaches, exemplified by DeepSurv \cite{katzman2018deepsurv}, integrate neural networks to capture nonlinear relationships while preserving the structure of models such as the Cox proportional hazards model.
Nonparametric approaches, such as DeepHit \cite{lee2018deephit} and CQRNN \cite{pearce2022censored}, leverage deep learning to directly estimate survival functions without relying on traditional assumptions.
These advances allow survival models to handle complex, high-dimensional data with greater precision and flexibility.

Naturally, each approach has limitations that may affect its suitability for different applications.
Parametric models rely on strong assumptions about the underlying distribution, which may not accurately capture true survival patterns.
Semiparametric models are dependent on the proportional hazards assumption, which can be invalid in certain datasets.
Nonparametric models, such as DeepHit and CQRNN, tend to be computationally intensive and require large datasets for effective training, making them less practical in resource-constrained settings.
Additionally, these models often produce discrete estimates, which may compromise interpretation and summarization flexibility compared to the continuous modeling offered predominantly by parametric models.
To address these limitations, we propose a parametric survival analysis method based on the Asymmetric Laplace Distribution (ALD).
Our contributions are listed below.
\begin{itemize}[topsep=0mm,itemsep=0mm,leftmargin=3mm]
    \item We introduce a flexible parametric survival model based on the Asymmetric Laplace Distribution, which offers superior flexibility in capturing diverse survival patterns compared to other distributions (parametric methods).
    \item The continuous nature of the ALD-based approach offers great flexibility in summarizing distribution-based predictions, thus addressing the limitations of existing discretized nonparametric methods.
    \item Experiments on 14 synthetic datasets and 7 real-world datasets in terms of 9 performance metrics demonstrate that our proposed framework consistently outperforms both parametric and nonparametric approaches in terms of both discrimination and calibration.
    These results underscore the robust performance and generalizability of our method in diverse datasets.
\end{itemize}

\section{Background}
\textbf{Survival Data.} A survival dataset $\mathcal{D}$ is represented as a set of triplets $\{(\mathbf{x}_n, y_n, e_n)\}_{n=1}^{N}$, where $\mathbf{x}_n \in \mathbb{R}^d$ denotes the set of covariates in $d$ dimensions, $y_n = \mathrm{min}(o_n, c_n) \in \mathbb{R}_+$ represents the observed time, and $e_n$ is the event indicator.
If the event of interest is observed, {\em e.g.} death, then $o_n < c_n$ and the event indicator is set to $e_n = 1$, otherwise, the event is {\em censored} and $e_n = 0$.
In this work, we make the common assumption that the distributions of observed and censored variables are conditionally independent given the covariates, {\em i.e.}, $ o \perp\!\!\!\!\perp c \mid \mathbf{x} $.
Moreover, while we primarily consider right-censored data, less common types of censoring can be readily implemented \cite{klein2006survival}, {\em e.g.}, left-censoring can be data handled by changing the likelihood accordingly (see Section~\ref{sec:ald_loss} for an example of how the maximum likelihood loss proposed here can be adapted for such a case).

\textbf{Survival and Hazard Functions.} Survival and hazards functions are two fundamental concepts in survival analysis.
The survival function is denoted as \( S(t) = P(T > t) \), which represents the probability that an individual has \emph{survived} beyond time \( t \). It can also be expressed in terms of the cumulative distribution function (CDF), \( F(t) \), which gives the probability that the event has occurred by the time \( t \), as $S(t) = 1 - F(t)$.
The hazards function, denoted as \( \lambda(t) \), describes the instantaneous risk that the event occurs at a specific time \( t \), given that the individual has survived up to that point.
Formally, it is defined as:
\[
\lambda(t) = \lim_{\Delta t \to 0} \frac{P(t \leq T < t + \Delta t | T \geq t)}{\Delta t} \,.
\]
The hazards function is related to the survival function through:
\[
\lambda(t) = -\frac{d}{dt} \log S(t) \,, \ \ {\rm or} \ \ 
S(t) = \exp\left(-\int_0^t \lambda(u) \, du\right).
\]
Furthermore, the probability density function (PDF), \( f(t) \), which represents the likelihood that the event occurs at time \( t \), can be derived as:
\[
f(t) = -\frac{d}{dt} S(t) = \lambda(t) S(t).
\]
These relationships establish a unified framework linking \( S(t) \), \( F(t) \), \( \lambda(t) \), and \( f(t) \), highlighting their interdependence in survival analysis.
Importantly, for the purpose of making predictions, we are interested in distributions {\em conditioned} on observed covariates, namely $S(t | \mathbf{x})$, $F(t | \mathbf{x})$, $\lambda(t | \mathbf{x})$ and $f(t | \mathbf{x})$.

\textbf{Survival Models.} Survival models can be broadly classified into three main categories. 
Parametric models assume that the survival PDF follows a specific probability distribution as descrived above.
These models thus use a predefined closed-form distribution to describe $f(t | \mathbf{x})$ and $F(t | \mathbf{x})$, for which a model estimating its parameters can be specified. 
Alternatively, semiparametric models, such as the Cox proportional hazards model \cite{cox1972regression}, first decompose the conditional hazards function as $\lambda(t \mid \mathbf{x})=\lambda(t)\lambda(\mathbf{x})$, then estimate $\lambda(t)$ from the data and specify a model for $\lambda(\mathbf{x})$. 
In contrast, nonparametric models, such as DeepHit and CQRNN \cite{pearce2022censored} circumvent directly modeling conditional distributions by discretizing $f(t | \mathbf{x})$ \citep[DeepHit,][]{lee2018deephit}, learning summaries of $f(t | \mathbf{x})$ such as (a fixed set of) quantiles \citet[CQRNN,][]{pearce2022censored}, or even learning to sample from $f(t | \mathbf{x})$ \cite{chapfuwa2018adversarial}.
More details can be found in Appendix~\ref{appendix:a2}. 

\section{Methods}
\subsection{Asymmetric Laplace Distribution (ALD)}
\begin{definition}[\bf\citet{kotz2012laplace}]\label{cor: 1}
A random variable $Y$ is said to have an asymmetric Laplace distribution
with parameters $(\theta, \sigma, \kappa)$, if its PDF is:
\begin{equation}\label{eq:ald_f}
\resizebox{0.91\hsize}{!}{$
f_{\text{ALD}}(y; \theta, \sigma, \kappa) = \frac{\sqrt{2}}{\sigma} \frac{\kappa}{1 + \kappa^2} \left\{ 
\begin{array}{ll}
\exp\left(\frac{\sqrt{2}\kappa}{\sigma}(\theta - y)\right), & \text{if } y \geq \theta \,, \\[12pt]
\exp\left(\frac{\sqrt{2}}{\sigma \kappa}(y - \theta)\right), & \text{if } y < \theta \,,
\end{array} \right.
$}
\end{equation}
where $\theta$, $\sigma > 0$, and $\kappa > 0$, are the location, scale and asymmetry parameters.
\end{definition}
Moreover, its CDF can be expressed as:
\begin{equation}\label{eq:ald_F}
\resizebox{0.91\hsize}{!}{$
F_{\text{ALD}}(y; \theta, \sigma, \kappa) =
\begin{cases} 
1 - \frac{1}{1 + \kappa^2} \exp\left( \frac{\sqrt{2}\kappa}{\sigma} (\theta - y) \right), & \text{if } y \geq \theta \,, \\[12pt]
\frac{\kappa^2}{1 + \kappa^2} \exp\left( \frac{\sqrt{2}}{\sigma \kappa} (y - \theta) \right), & \text{if } y < \theta \,.
\end{cases}
$}
\end{equation}
We denote the distribution of $Y$ as $\mathcal{AL}(y; \theta, \sigma, \kappa)$.
\begin{corollary} \label{cor: 2}
The Asymmetric Laplace Distribution, denoted as $\mathcal{AL}(\theta, \sigma, \kappa)$, can be reparameterized as $\mathcal{AL}(\theta, \sigma, q)$ to facilitate quantile regression \cite{yu2001bayesian}, where $q \in (0, 1)$ is the percentile parameter that represents the desired quantile. The relationship between $q$ and $\kappa$ is given by $q = \kappa^2/(\kappa^2 + 1)$.
\end{corollary}
Additional details are provided in Appendix \ref{appendix:a1}.

\begin{figure}[t] 
    \centering
    \includegraphics[width=\linewidth]{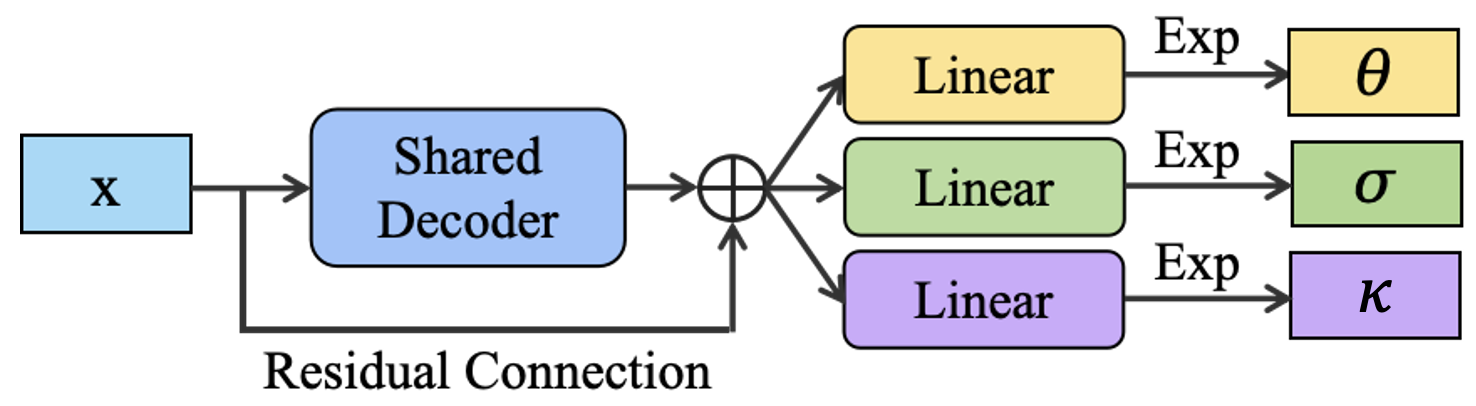} 
    \vspace{-7mm}
    \caption{The proposed neural network architecture for predicting the parameters of the Asymmetric Laplace Distribution $\mathcal{AL}(\theta, \sigma, \kappa)$.}
    \label{fig:network_architecture} 
\end{figure}

\subsection{Model for the ALD}
The structure of the proposed model is illustrated in Figure~\ref{fig:network_architecture}, where a shared encoder is followed by three independent heads to estimate the parameters $\theta$, $\sigma$, and $\kappa$ of the ALD distribution.
For the purpose of the experiments in Section~\ref{sec:experiments} with structured data, we use fully connected layers with ReLU activation functions.
The outputs of the model connected to $\theta$, $\sigma$ and $\kappa$ are further constrained to be non-negative through an exponential (Exp) activation. 
In addition, a residual connection is included to enhance gradient flow and improve model stability.
See Appendix \ref{appendix:b3} for more details about the architecture of the model.

\subsection{Learning for the ALD}\label{sec:ald_loss}
We propose learning the model for the ALD through maximum likelihood estimation (MLE).
Since the event of interest can be either observed or censored, we specify separate objectives for these two types of data.
For observed events, for which $e=1$, we directly seek to optimize the parameters of the model to maximize \(f_{\text{ALD}}(t | \mathbf{x})\) in \eqref{eq:ald_f}.
Alternatively, for censored events, for which $e=0$, we optimize the parameters of the model to maximize the survival function \(S_{\text{ALD}}(t | \mathbf{x})=1-F_{\text{ALD}}(t | \mathbf{x})\) in \eqref{eq:ald_F}.
In this manner, the ALD objective below accounts for both the occurrence of events and their respective timing while explicitly incorporating the survival probability constraint for censored data:
\begin{align}\label{eq:ald_loss}
- \mathcal{L}_{\text{ALD}} & = \sum_{n \in \mathcal{D}_{\text{O}}} \log f_{\text{ALD}}(y_n \mid \mathbf{x}_n) \nonumber\\
& + \sum_{n \in \mathcal{D}_{\text{C}}} \log S_{\text{ALD}}(y_n \mid \mathbf{x}_n) \,,
\end{align}
where $\mathcal{D}_{\text{O}}$ and $\mathcal{D}_{\text{C}}$ are the subsets of $\mathcal{D}=\mathcal{D}_{\text{O}}\cup\mathcal{D}_{\text{C}}$ for which $e=1$ and $e=0$, respectively.
Detailed derivations of the objective in \eqref{eq:ald_loss} be found in Appendix \ref{appendix:a1}.

The simplicity of the objective in \eqref{eq:ald_loss} is a consequence of the ability to write the relevant distributions, \(f_{\text{ALD}}(t | \mathbf{x})\) and \(S_{\text{ALD}}(t | \mathbf{x})\), in closed form.
Moreover, we make the following remarks.
\begin{itemize}[topsep=0mm,itemsep=0mm,leftmargin=3mm]
\item The objective in \eqref{eq:ald_loss} has the same form as the one used in other parametric approaches, for instance \citet{royston2001lognormal} for the log-normal distribution.
\item We can readily adapt the loss for other forms of censoring, for instance, if events are left censored, we only have to replace the second term of \eqref{eq:ald_loss} by \(1-S_{\text{ALD}}(t | \mathbf{x})\).
\item We do not consider additional loss terms, as is usually done for other approaches, {\em e.g.}, DeepHit optimizes a form similar to \eqref{eq:ald_loss}, where the density function and cumulative distribution are replaced by discretized approximations, but also consider an additional loss term to improve discrimination \cite{lee2018deephit}.
\item Although the ALD in \eqref{eq:ald_f} has support for $t<0$, we have observed empirically that this is unlikely to happen, as we will demonstrate in the experiments.
\end{itemize}

\subsection{Comparison between our Method and CQRNN}
CQRNN \cite{pearce2022censored} adopts the widely-used objective for quantile regression, which is also based on the Asymmetric Laplace Distribution $\mathcal{AL}(\theta, \sigma, q)$, and uses the transformation in Corollary \ref{cor: 2}.
Specifically, they use the maximum likelihood estimation approach to optimize the following objective:
\begin{align}
\mathcal{L}_{\text{QR}}(y;\theta_q, \sigma, q) & = \log \sigma - \log [q(1-q)] \nonumber\\
& + \frac{1}{\sigma}
\begin{cases} 
q (y - \theta_q), & \text{if } y \geq \theta_q \,, \\
(1-q)(\theta_q - y), & \text{if } y < \theta_q \,.
\end{cases} \label{eq:cqrnn_loss}
\end{align}
Following the quantile regression framework, their approach optimizes a model to predict $\theta_q$ for a predefined collection of quantile values, {\em e.g.}, $q=\{0.1,0.2,\ldots,0.9\}$.  
Effectively and similarly to ours, \citet{pearce2022censored} specify a shared encoder with multiple heads to predict $\{\theta_q\}_q$.
Note that the objective in \eqref{eq:cqrnn_loss} does not require one to specify $\sigma$, which results in the following simplified loss:
\begin{align}\label{eq:pinball_loss}
\mathcal{L}_{\text{QR}}(y; \theta_q, q) & =  
\begin{cases} 
q (y - \theta_q), & \text{if } y \geq \theta_q \,, \\
(1 - q)(\theta_q - y), & \text{if } y < \theta_q \,,
\end{cases} \nonumber\\
& = \ (y-\theta_q)(q - \mathbb{I}[\theta_q > y]) \,,
\end{align}
where $\mathbb{I}[\cdot]$ is the indicator function.
The formulation in \eqref{eq:pinball_loss} is also known as the {\em pinball} or {\em checkmark} loss \cite{koenker1978regression}, which is widely used in the quantile regression literature.

Importantly, unlike in the objective for our approach in \eqref{eq:ald_loss}, CQRNN does not maximize the survival probability directly.
Instead, they adopt the also widely used approach based on the Portnoy’s estimator \cite{neocleous2006correction}, which optimizes an objective function tailored for censored quantile regression.
Specifically, this approach introduces a re-weighting scheme to handle the censored data:
\begin{align}\label{eq:cqr_loss}
\mathcal{L}_{\text{CQR}}(y, y^*; \theta_q, q, w) & = w \mathcal{L}_{\text{QR}}(y; \theta_q, q) \nonumber\\
& + (1-w)\mathcal{L}_{\text{QR}}(y^*; \theta_q, q) \,,
\end{align}
where where $y^*$ is a {\em pseudo} value set to be ``sufficiently'' larger than all the observed values of $y$ in the data. 
Specifically, in CQRNN \cite{pearce2022censored} it is defined as $y^* = 1.2 \max_i y_i$. 
However, Portnoy \cite{neocleous2006correction} indicates that $y^*$ could be set to any sufficiently large value approximating $\infty$. 
For example, \citet{koenker2022quantreg} sets $y^*=1e6$. 
This means that in practice, this parameter often requires careful tuning based on the specific dataset, provided that different datasets exhibit varying levels of sensitivity to it. 
In some cases, we have observed that small perturbations in $y^*$ can lead to considerable variation on performance metrics. 
Consequently, optimizing this parameter can be non-trivial, making the use of CQRNN, and other censored quantile regression methods, challenging.

The other parameter in \eqref{eq:cqr_loss} that requires attention is the weight $w\in(0,1)$, which is defined as $w = (q - q_c)/(1 - q_c)$, and where $q_c$ is the quantile at which the data point was censored ($e=0, y=c$), with respect to the observed value distribution, {\em i.e.}, $p(o < c|x)$.
The challenge is that $q_c$ is not known in practice.
%
To address this issue, CQRNN proposes two strategies: a sequential grid algorithm and the quantile grid output algorithm. 
The core idea of both strategies is to approximate $q_c$ using the proportion $q$ corresponding to the quantile that is closest to the censoring value $c$ using the distribution of observed events $y$, which are readily available.
Even with this approach, $q_c$ is an inherently inaccurate approximation. 
Its precision heavily depends on the initial grid of $q$ values, specifically, the intervals between consecutive $q$ values. 
Consequently, smaller intervals provide finer granularity, but increased computational costs, while larger intervals may lead to coarser approximations that tend to affect model performance.
This means that in some cases, the model is sensitive to the choice of the grid of $q$ values.

In contrast, our approach enjoys a simple objective function resulting in parametric estimates of several distribution summaries such as mean, median, standard deviation, and quantiles without additional cost.
Additional details of CQRNN are provided for completeness in Appendix~\ref{appendix:a2}.

\section{Related Work}
Survival analysis is a fundamental area of study in statistics and machine learning, focusing on modeling time-to-event data while accounting for censoring.
A wide range of models has been developed that span parametric, semiparametric, and nonparametric methods.

Parametric models assume a specific distribution for the time-to-event variable, providing a structured approach to modeling survival and hazards functions.
Commonly used distributions include the exponential \cite{feigl1965estimation}, Weibull \cite{scholz1996maximum}, and the log-normal distribution \cite{royston2001lognormal}.
For example, the log-normal model assumes that the logarithm of survival times follows a normal distribution, enabling straightforward parameterization of survival curves.
In modern approaches \cite{hoseini2017comparison}, neural networks are employed to learn the parameters of the assumed distribution, {\em e.g.}, the mean and variance for the log-normal.
This combination allows the model to leverage the power of neural networks to capture complex, nonlinear relationships between covariates and survival times, while keeping the interpretability and structure inherent to the parametric framework.
However, these models face challenges despite their simplicity when the true event distribution significantly deviates from that assumed.

Semiparametric methods strike a balance between flexibility and interpretability.
One notable example is the Cox proportional hazards model \cite{cox1972regression}, which assumes a multiplicative effect of covariates on the hazard function.
Building on this foundation, DeepSurv \cite{katzman2018deepsurv}, a deep learning-based extension, replaces the linear assumption with neural network architectures to model complex feature interactions.
DeepSurv has demonstrated improved performance in handling high-dimensional covariates while maintaining the interpretability of hazard ratios.
However, semiparametric models face challenges in effectively handling censored data, particularly when censoring rates are very high.
In such cases, the limited amount of usable information can lead to degraded performance and reduced reliability of the model's estimates.

The Kaplan-Meier (KM) estimator \cite{kaplan1958nonparametric} is a widely used nonparametric method for survival analysis.
It estimates the survival function directly from the data without assuming any underlying distribution.
The KM estimator is particularly effective for visualizing survival curves and computing survival probabilities.
However, its inability to incorporate covariates limits its applicability in complex scenarios.
More recent nonparametric approaches, such as DeepHit \cite{lee2018deephit} and CQRNN \cite{pearce2022censored}, leverage neural networks to predict survival probabilities or quantiles without imposing strong distributional assumptions.
These methods are highly flexible, capturing nonlinear relationships between features and survival outcomes, making them particularly suited for high-dimensional and heterogeneous datasets.
Nevertheless, a notable shortcoming of both DeepHit and CQRNN is that they produce piecewise constant or point-mass distribution estimates, respectively, that lack continuity and smoothness, leading to survival estimates that can complicate summarization, interpretation, and downstream analysis.

\begin{table}[t]
\caption{Dataset summaries: number of features (Feats), training/test data size, and proportion of censored events (PropCens).}
\label{tab:dataset}
\vspace{1mm}
\resizebox{\columnwidth}{!}{
\begin{tabular}{ccccc}
\toprule
\textbf{Dataset} & \textbf{Feats} & \textbf{Train data} & \textbf{Test data} & \textbf{PropCens} \\
\midrule
\multicolumn{5}{c}{\textbf{Type 1 -- Synthetic target data with synthetic censoring}} \\
Norm linear       & 1 & 500  & 1000 & 0.20 \\
Norm non-linear   & 1 & 500  & 1000 & 0.24 \\
Exponential       & 1 & 500  & 1000 & 0.30 \\
Weibull           & 1 & 500  & 1000 & 0.22 \\
LogNorm           & 1 & 500  & 1000 & 0.21 \\
Norm uniform      & 1 & 500  & 1000 & 0.62 \\
Norm heavy        & 4 & 2000 & 1000 & 0.80 \\
Norm med          & 4 & 2000 & 1000 & 0.49 \\
Norm light        & 4 & 2000 & 1000 & 0.25 \\
Norm same         & 4 & 2000 & 1000 & 0.50 \\
LogNorm heavy     & 8 & 4000 & 1000 & 0.75 \\
LogNorm med       & 8 & 4000 & 1000 & 0.52 \\
LogNorm light     & 8 & 4000 & 1000 & 0.23 \\
LogNorm same      & 8 & 4000 & 1000 & 0.50 \\
\midrule
\multicolumn{5}{c}{\textbf{Type 2 -- Real target data with real censoring}} \\
METABRIC          & 9 & 1523 & 381  & 0.42 \\
WHAS              & 6 & 1310 & 328  & 0.57 \\
SUPPORT           & 14 & 7098 & 1775 & 0.32 \\
GBSG              & 7 & 1785 & 447  & 0.42 \\
TMBImmuno         & 3 & 1328 & 332  & 0.49 \\
BreastMSK         & 5 & 1467 & 367  & 0.77 \\
LGGGBM            & 5 & 510  & 128  & 0.60 \\
\bottomrule
\end{tabular}
}
\vspace{-6mm}
\end{table}

\section{Experiments}\label{sec:experiments}
\subsection{Datasets}
We utilize two types of datasets, following \citet{pearce2022censored}: 
(Type 1) synthetic data with synthetic censoring and
(Type 2) real-world data with real censoring. 
Table~\ref{tab:dataset} presents a summary of general statistics for all datasets.
To account for training and model initialization variability, we run all experiments 10 times with random splits of the data with partitions consistent with Table~\ref{tab:dataset}.
The source code required to reproduce the experiments presented in the following can be found in the Supplementary Material.

For synthetic observed data with synthetic censoring, the input features $\mathbf{x}$ are generated uniformly as $\mathbf{x} \sim \mathcal{U} (0, 2)^d$, where $d$ represents the number of features. 
The observed variable $o \sim p(o | \mathbf{x})$ and the censored variable $c \sim p(c | \mathbf{x})$ follow distinct distributions, with each distribution parameterized differently, depending on the specific dataset configuration. 
This variability in distributions and parameters allows for the evaluation of the model's robustness under diverse synthetic data scenarios. 

For real target data with real censoring, we utilize datasets that span various domains, characterized by distinct features, sample sizes, and censoring proportions. 
Four of these datasets: METABRIC, WHAS, SUPPORT, and GBSG, were retrieved from the DeepSurv GitHub repository\footnote{ https://github.com/jaredleekatzman/DeepSurv/}.
Other details are available in \citet{katzman2018deepsurv}. 
The remaining three datasets: TMBImmuno, BreastMSK, and LGGGBM were sourced from cBioPortal\footnote{https://www.cbioportal.org/} for Cancer Genomics.
These datasets constitute a diverse benchmark across domains such as oncology and cardiology, allowing a comprehensive evaluation of survival analysis methods.
Additional details of all datasets can be found in Appendix~\ref{appendix:b1}.

\subsection{Metrics}
\textbf{Predictive Accuracy Metrics}: Mean Absolute Error (MAE) and Integrated Brier Score (IBS) \cite{graf1999assessment}, measure the accuracy of survival time predictions. 
MAE quantifies the average magnitude of errors between predicted and observed survival times $\tilde{y}_i$ and $y_i$, respectively. 
For synthetic data, ground truth values are obtained directly from the observed distribution, while for real data, only observed events ($e = 1$) are considered. 
For the IBS calculation, we select 100 time points evenly from the 0.1 to 0.9 quantiles of the distribution for $y$ in the training set.

\textbf{Concordance Metrics}: Harrell’s C-Index \cite{harrell1982evaluating} and Uno’s C-Index \cite{uno2011c}, which evaluate the ability of the model to correctly order survival times in a pairwise manner, while accounting for censoring.
Harrell’s C-Index is known to be susceptible to bias, when the censoring rate is high.
This happens because censoring dominates the pairwise ranking when estimating the proportion of correctly ordered event pairs.
Alternatively, Uno’s C-Index adjusts for censoring by using inverse probability weighting, which provides a more robust estimate when the proportion of censored events is high.

\begin{table*}[t]
\caption{Summary of benchmarking results across 21 datasets. Each column group shows three figures: the number of datasets where our method significantly outperforms, underperforms or is comparable with the baseline indicated.
The last two rows summarize the column totals and proportions to simplify the comparisons.
For reference, the total number of comparisons is 189.
}
\label{tab:comparison}
\vspace{1mm}
\resizebox{\textwidth}{!}{
\begin{tabular}{ccccccccccccc}
\toprule
\multirow{2}{*}{\textbf{Metric}}   & \multicolumn{3}{c}{\textbf{{\em vs}. CQRNN}} & \multicolumn{3}{c}{\textbf{{\em vs}. LogNorm}} & \multicolumn{3}{c}{\textbf{{\em vs}. DeepSurv}} & \multicolumn{3}{c}{\textbf{{\em vs}. DeepHit}} \\ 
                            & \textbf{Better} & \textbf{Worse} & \textbf{Same} & \textbf{Better} & \textbf{Worse} & \textbf{Same} & \textbf{Better} & \textbf{Worse} & \textbf{Same} & \textbf{Better} & \textbf{Worse} & \textbf{Same} \\ \midrule
\textbf{MAE}                & 6  & 8 & 7    & 10 & 3 & 8    & 6  & 8  & 7    & 12 & 6 & 3 \\
\textbf{IBS}                & 19 & 1 & 1    & 21 & 0 & 0    & 21 & 0  & 0    & 21 & 0 & 0 \\
\textbf{Harrell’s C-Index}  & 4  & 2 & 15   & 10 & 3 & 8    & 6  & 2  & 13   & 15 & 0 & 6 \\
\textbf{Uno’s C-Index}      & 2  & 3 & 16   & 9  & 2 & 10   & 6  & 1  & 14   & 15 & 0 & 6 \\
\textbf{CensDcal}           & 8  & 4 & 9    & 10 & 1 & 10   & 8  & 5  & 8    & 15 & 1 & 5 \\
\textbf{Cal $[S(t|\mathbf{x})]$(Slope)}             & 0  & 0 & 21   & 15 & 0 & 6    & 13 & 0  & 8    & 12 & 0 & 9 \\
\textbf{Cal $[S(t|\mathbf{x})]$(Intercept)}         & 0  & 0 & 21   & 14 & 0 & 7    & 0  & 11 & 10   & 16 & 0 & 5 \\
\textbf{Cal $[f(t|\mathbf{x})]$(Slope)}             & 4  & 0 & 17   & 14 & 0 & 7    & 9  & 0  & 12   & 14 & 0 & 7 \\
\textbf{Cal $[f(t|\mathbf{x})]$(Intercept)}         & 0  & 4 & 17   & 10 & 0 & 11   & 8  & 0  & 13   & 18 & 0 & 3 \\ 
\midrule
\textbf{Total}              & 43 / 189 & 22 / 189 & 124 / 189 & 113 / 189 & 9 / 189 & 67 / 189  & 77 / 189 & 27 / 189 & 85 / 189 & 138 / 189 & 7 / 189 & 44 / 189  \\
\textbf{Proportion}              & 0.228 & 0.116 & 0.656 & 0.598 & 0.048 & 0.354  & 0.407 & 0.143 & 0.450 & 0.730 & 0.037 & 0.233 \\ \bottomrule
\end{tabular}
}
\\
\\
\\
\end{table*}

\textbf{Calibration Metrics}: There are several metrics to assess calibration.
We consider summaries (slope and intercept) of the calibration curves using the predicted PDF $f(t | \mathbf{x})$ or the survival distribution $S(t | \mathbf{x})$.
Moreover, we use the censored D-Calibration (CensDcal) \cite{haider2020effective}.
For the former, Cal$[f(t \mid \mathbf{x})]$, 9 prediction interval widths are considered, {\em e.g.}, 0.1 for $[0.45,0.55]$, 0.2 for $[0.4, 0.6]$, {\em etc}.
These are used to define the time ranges for each prediction, after which we calculate the proportion of test events that fall within each interval.
The calculation of the proportion of censored and observed cases follows the methodology in \citet{goldstein2020x}, with further details provided in Appendix \ref{appendix:b2}.
This calibration curve of expected {\em vs}. observed events is summarized with an ordinary least squares linear fit parameterized by its {\em slope} and {\em intercept}.
A well-calibrated model is expected to have a unit slope and a zero intercept.
For the survival distribution, Cal$[S(t | \mathbf{x})]$, we follow a similar procedure, however, we consider 10 non-overlapping intervals in the range $(0,1)$, {\em i.e.}, $(0,0.1]$, $(0.1,0.2]$, {\em etc} and then calculating the proportion of test events that fall within each interval.
The calculation of CensDcal starts with that of Cal$[S(t | \mathbf{x})]$, which is followed by computing the sum of squared residuals between the observed and expected proportions, {\em i.e.}, 0.1 for the 10 intervals defined above.

These three groups of metrics provide a robust framework for evaluating predictive accuracy, calibration, and concordance in survival analysis.
For the results we calculate averages and standard deviations for all metrics over 10 random test sets.
The metrics that require a point estimate, {\em i.e.}, MAE and C-Index are obtained using the expected value of $f(t | \mathbf{x})$, which can be calculated in closed form.
More details about all metrics can be found in Appendix \ref{appendix:b2}.

\begin{figure*}[t]
    \centering
    \subfigure[Concordance metric (Harrell's C-index).]{
        \includegraphics[width=1\linewidth]{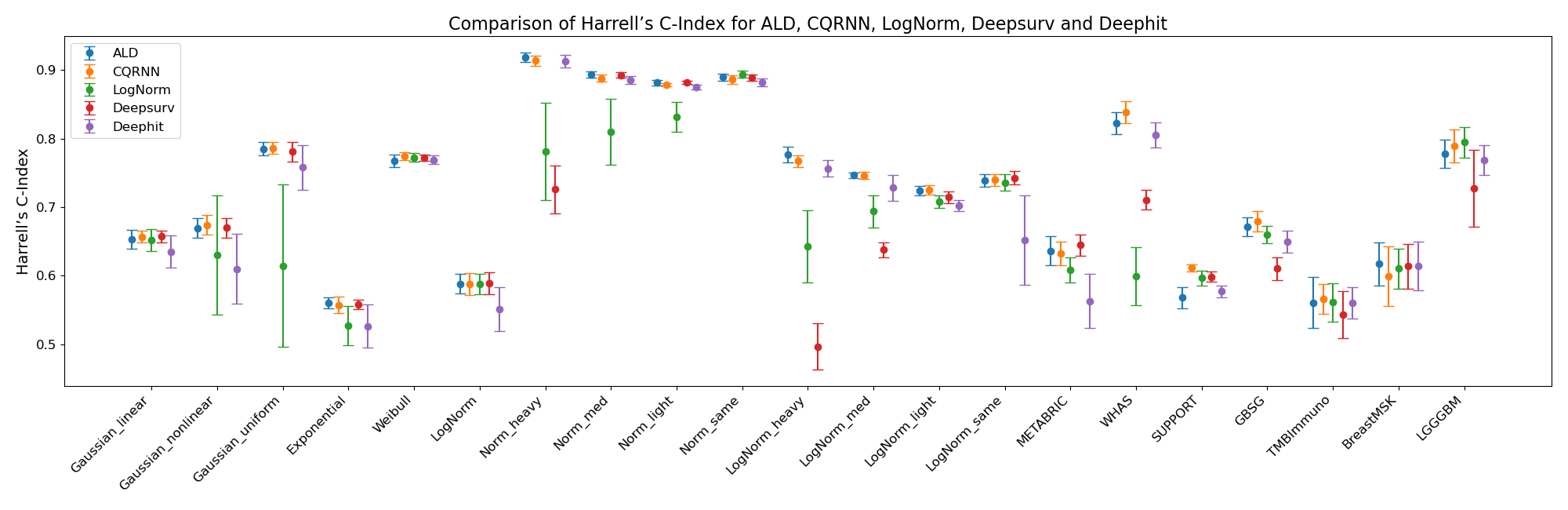}
        \label{fig:c_index}
    }
    \vspace{-2mm}
    \subfigure[Calibration metric (CensDcal).]{
        \includegraphics[width=1\linewidth]{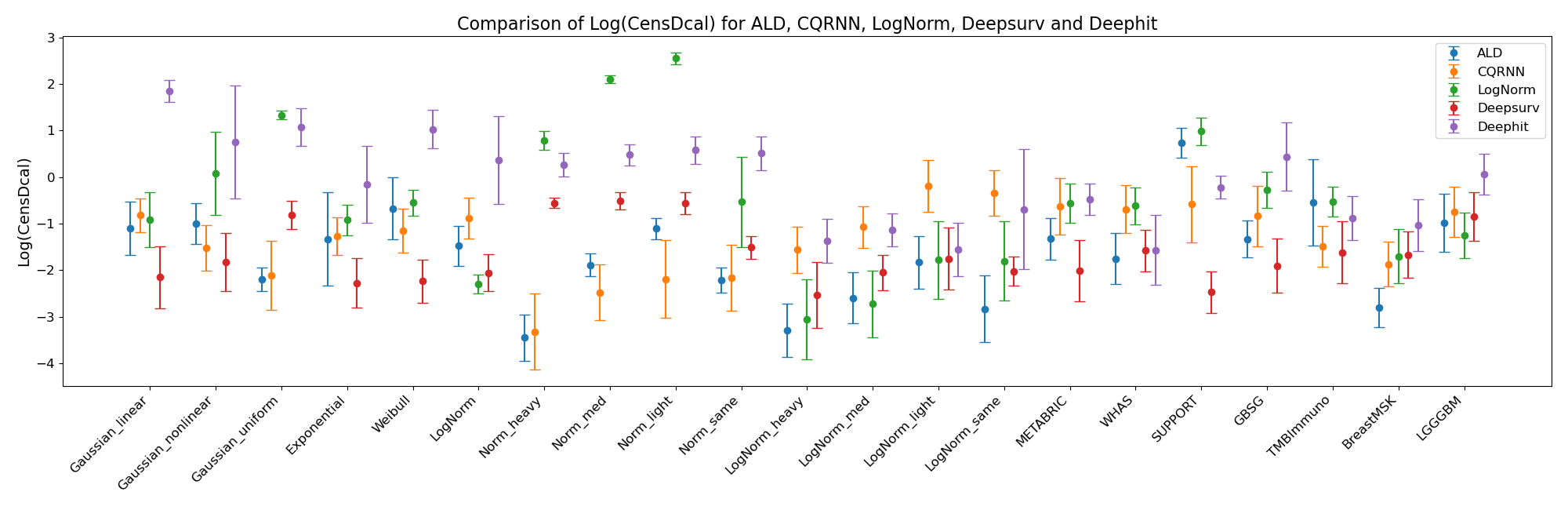}
        \label{fig:dcal_cens}
    }
    \vspace{-1mm}
    \caption{Performance on discrimination and calibration metrics. (a) concordance and (b) calibration. Reported are test averages with standard deviations over 10 runs.}
    \label{fig:performance}
\end{figure*}

\subsection{Baselines}
We compare the proposed method against four baselines representative of the related work, to evaluate performance and effectiveness.
\textbf{LogNorm} \cite{royston2001lognormal}: A parametric survival model that assumes that the event times follow a log-normal distribution. 
\textbf{DeepSurv} \cite{katzman2018deepsurv}: A semi-parametric survival model based on the Cox proportional hazards framework, leveraging deep neural networks for the representation of the time-independent hazards. 
\textbf{DeepHit} \cite{lee2018deephit}: A deep learning-based survival model that predicts piece-wise probability distributions over event times using a fully neural network architecture. 
\textbf{CQRNN} \cite{pearce2022censored}: A censored quantile regression model that employs a neural network architecture, and whose objective is based on the Asymmetric Laplace Distribution.
These baselines represent a mix of parametric, semi-parametric, and non-parametric survival modeling techniques, allowing us to provide a comprehensive benchmark for comparison.
The implementation details, including model selection, of our method and the other baselines can be found in Appendix~\ref{appendix:b3}.

\subsection{Results}\label{sec:results}
Table~\ref{tab:comparison} provides a comprehensive summary of the comparisons between our model and the four baselines in 21 datasets and 9 evaluation metrics, which is 189 comparisons in total.
When assessing the statistical significance of the different metrics we use a Student's $t$ test with $p<0.05$ considered significant after correction for false discovery rate using Benjamini-Hochberg \cite{benjamini1995controlling}.
These results underscore several key insights:

\textbf{Overall Superiority}: Our model is significantly better than the baselines consistently more often.
For instance, our model significantly outperforms CQRNN in 23\% of the comparisons while the opposite only occurs 12\%, and these proportions are higher for the comparisons against the other baselines, namely, 60\%, 41\% and 73\% for LogNorm, DeepSurv and DeepHit, respectively.

\textbf{Accuracy}: Our model demonstrates significant improvements over the baselines in predictive accuracy, with a notable improvement in MAE compared to LogNorm and DeepHit. 
Moreover, it consistently outperforms the baselines on nearly every dataset when evaluated with the IBS metric. 
This consistent superiority in IBS underscores our model’s ability to provide accurate and reliable predictions across the entire time range, not just at specific time points.
Table~\ref{tab:overall results} and Figure~\ref{fig:overall_results} in the Appendix further support this, showing that our method achieves significantly lower IBS values, which reflects its effectiveness in learning from censored data without exacerbating bias in survival estimates.

\textbf{Concordance}: While Harrell's and Uno’s C-Indices yield more balanced results across models, our model achieves a relatively higher number of wins compared to other baselines, especially in terms of Harrel's C and against LogNorm and DeepHit, which underscores our model's ability to rank predictions effectively.

\textbf{Calibration}: Our model demonstrates strong performance in calibration metrics, particularly in CensDcal, Cal$[S(t|\mathbf{x})]$ (slope) and Cal$[f(t|\mathbf{x})]$ (slope). 
The results for CensDcal highlight its ability to effectively handle censored observations. 
Additionally, our model shows significant superiority in slope- and intercept-related metrics, particularly when compared to log-norm and DeepHit. 
However, the improvement over CQRNN remains relatively subtle. 
These results indicate that our model achieves better calibration of predicted survival probabilities, ensuring closer alignment between predictions and observed outcomes in both the survival CDF $S(t|\mathbf{x})$ and PDF $f(t|\mathbf{x})$.
This underscores the reliability and robustness of our model in accurately capturing true survival behavior across diverse datasets.

\begin{figure}[t] 
    \centering
    \includegraphics[width=\linewidth]{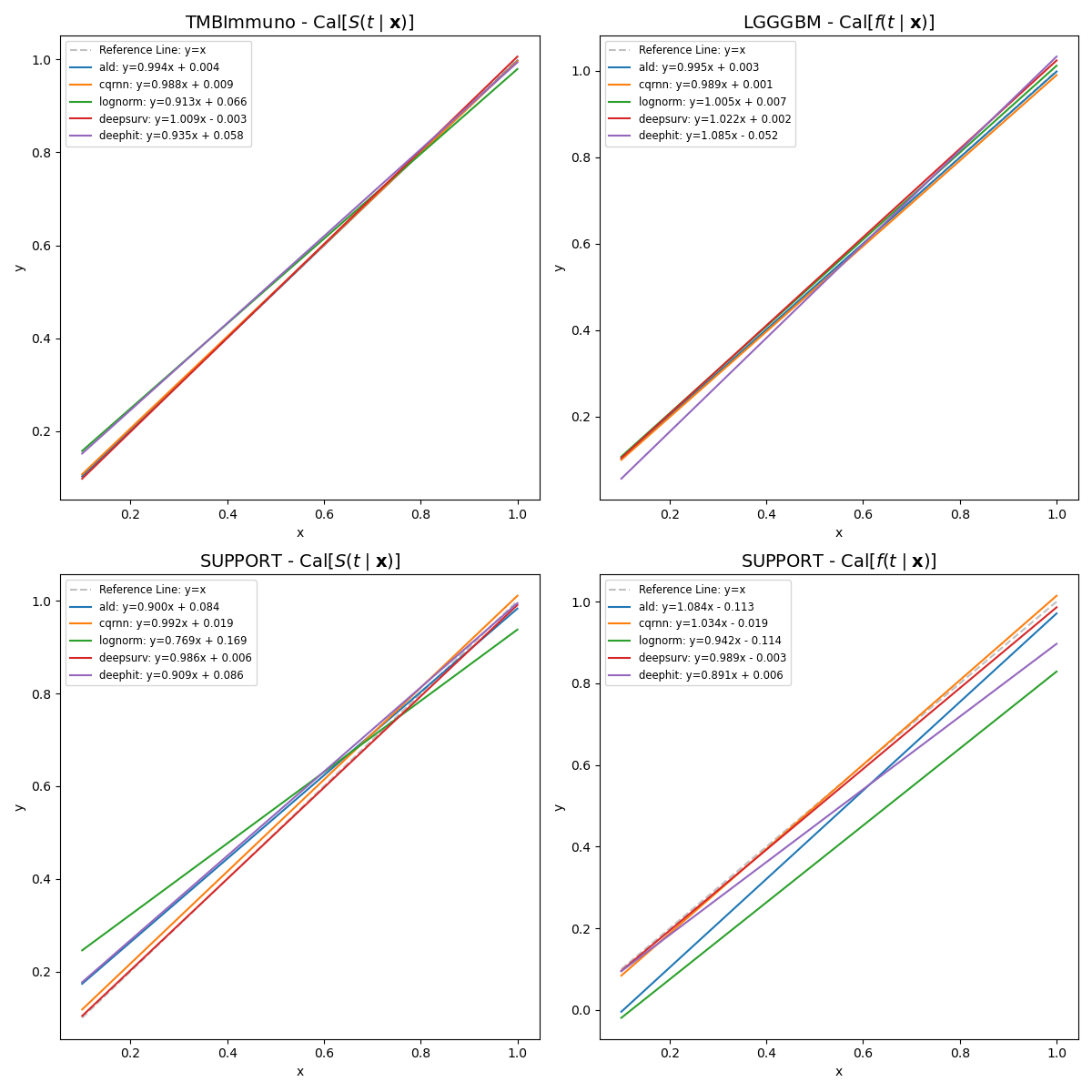} 
    \caption{Examples of best and worst calibration curves.
    Slope and intercept of the linear fit are shown in the legend.}
    \label{fig:lineplot} 
\end{figure}

For illustration purposes, Figure~\ref{fig:performance} shows the performance of all models and datasets using Harrell's C-index and CensDcal. 
Specifically, Figure~\ref{fig:c_index} provides a comparison of Harrell's C-index, highlighting the discriminative performance of our proposed model, ALD, alongside other baseline models.
The x axis lists all the datasets, both synthetic datasets ({\em e.g.}, Gaussian linear, exponential, {\em etc}.) and real-world datasets ({\em e.g.}, METABRIC, WHAS, {\em etc}.), while the y axis indicates the range of corresponding mean C-index values, with error bars representing standard deviation across 10 model runs.
Our method demonstrates consistently strong performance across both synthetic and real-world datasets, frequently achieving higher or comparable C-index values in relation to the baseline models.
Among these, CQRNN stands out as the most competitive alternative, achieving similar levels of performance on certain datasets.
However, in most cases, our model outperforms CQRNN, reflecting its robustness and superior discriminative capability.
In particular, ALD excels under scenarios with high censorsing rates.
For example, on Norm heavy (PropCens: 0.80), Norm med (PropCens: 0.49), LogNorm heavy (PropCens: 0.75), and LogNorm med (PropCens: 0.52), ALD consistently outperforms other models.
This performance highlights ALD's ability to effectively handle challenging scenarios.
Such robustness under high censorship further underscores ALD's reliability and adaptability in various survival analysis tasks.

Using a similar comparison framework, Figure~\ref{fig:dcal_cens} presents the calibration results using CensDcal.
Concisely, the proposed model achieves consistently better (lower) CensDcal figures across most datasets, reflecting superior calibration performance compared to the baseline models.

Complementary to the CensDcal calibration metric, the slope and intercept summaries of the calibration curve provide a more intuitive (and graphical) perspective of the calibration results.
Figure~\ref{fig:lineplot} presents the best (first row) and worst (second row) results from our model on real-world data.
The left and right columns represent the curves for Cal$[S(t | \mathbf{x})]$ and for Cal$[f(t | \mathbf{x})]$, respectively.
The gray dashed line represents the idealized result for which the slope is one and the intercept is zero.
%

The proposed model demonstrates exceptional performance on the TMBImmuno dataset for Cal$[f(S | \mathbf{x})]$ summaries, as well as on the LGGGBM dataset for Cal$[f(t | \mathbf{x})]$ summaries indicating robust calibration across both versions of the calibration metrics.
In contrast, the performance on the SUPPORT dataset is relatively weaker. 
This discrepancy can largely be attributed to our method's reliance on the assumption of the ALD, which may not be appropriate across all datasets.
This limitation is particularly evident in datasets like SUPPORT.
Notably, the SUPPORT data exhibit high skewness with a relatively small range of $y$. 
Specifically, Figure~\ref{fig:SUPPORT} in the Appendix shows the event distribution for the SUPPORT data, from which we can see that it is heavily skewed.
Such an skewness manifested as the concentration of events close to 0 makes it challenging to achieve good calibration in that range, {\em i.e.}, $t\to0$.
Similarly, our method attempted to predict smaller values for the initial quantiles but still allocated a disproportionately large weight to the first two intervals because of the small and highly concentrated predicted quantiles, which significantly reduced the capacity for the remaining intervals and ultimately degraded calibration performance.

However, the calibration results for this dataset remain within reasonable ranges and more importantly, comparable to those from the baselines. 
Detailed results for all datasets are provided in Appendix~\ref{appendix:c}. Overall and consistent with the summary results in Table~\ref{tab:comparison}, our model demonstrates a clear advantage on the slope and intercept metrics, consistently achieving better performance compared to the baselines.

%
Finally, we also explore other distribution summaries, {\em i.e.}, the mode and median, to evaluate their impact on the performance of MAE and C-index. 
Table~\ref{tab:overall results3} shows that different summaries may perform better on some datasets.
In addition, recognizing that the ALD has support for $t<0$, we summarized the empirical quantiles of the predicted $F_{\text{ALD}}(0 | \mathbf{x})$, {\em i.e.}, the probability that events occur up to $t=0$.
Interestingly, Table~\ref{tab:ald_cdf} in the Appendix indicate that this is not an issue most of the time, as in most cases, $F_{\text{ALD}}(0 | \mathbf{x})\to 0$ for most of the predictions by the model on the test set. 
%

\section{Conclusion}
In this paper, we proposed a parametric survival model based on the Asymmetric Laplace Distribution and provided a comprehensive comparison and analysis with existing methods, particularly CQRNN, which leverages the same distribution.
Our model produces closed-form distributions, which enables flexible summarization and interpretation of predictions.
Experimental results on a diverse range of synthetic and real-world datasets demonstrate that our approach offers very competitive performance in relation to multiple baselines across accuracy, concordance, and calibration metrics.

\textbf{Limitations.} First, our method relies on the assumption of the Asymmetric Laplace Distribution, which may not be universally applicable. This limitation was particularly evident in certain cases, such as with the SUPPORT dataset, as highlighted in Section~\ref{sec:results}, where the performance of our method faced challenges, especially in terms of calibration.
Second, while our approach facilitates the calculation of different distribution metrics such as mean, median, mode, and even distribution quantiles, selecting the most suitable summary statistic for specific datasets or applications remains a non-trivial task.
In this study, we selected the mean as the main summary statistic, which results in relatively balanced performance metrics; however, it does not offer an advantage for example, in terms of C-index and MAE when compared to CQRNN.
Nevertheless, considering other summary statistics as part of model selection, which we did not attempt, may improve performance on these metrics for certain datasets, as detailed in Appendix~\ref{appendix:c}.

\section*{Acknowledgements}
This work was supported by grant 1R61-NS120246-02 from the National Institute of Neurological Disorders and Diseases (NINDS).

\section*{Impact Statement}
Our proposed survival analysis method utilizes the Asymmetric Laplace Distribution (ALD) to deliver closed-form solutions for key event summaries, such as means and quantiles, facilitating more interpretable predictions. 
The method outperforms both traditional parametric and nonparametric approaches in terms of discrimination and calibration by optimizing individual-level parameters through maximum likelihood. 
This advancement has significant implications for applications like personalized medicine, where accurate and interpretable predictions of event timing are crucial.

\bibliography{example_paper}
\bibliographystyle{icml2025}

\newpage
\appendix
\onecolumn
\section{Analytical Results}
This section provides the analytical results. 
Detailed proofs for the Asymmetric Laplace Distribution Loss can be found in Appendix \ref{appendix:a1}, while the analysis of all the baselines, including CQRNN, LogNormal MLE, DeepSurv, and DeepHit, is presented in Appendix \ref{appendix:a2}.

\subsection{Proofs for the Asymmetric Laplace Distribution Loss}
\label{appendix:a1}
\textbf{Theorem 1.}  
If \( Y \sim \mathcal{AL}(\theta, \sigma, \kappa) \), where \( \mathcal{AL} \) denotes the Asymmetric Laplace Distribution with location parameter \( \theta \), scale parameter \( \sigma > 0 \), and asymmetry parameter \( \kappa > 0 \), then the ALD loss is given by:

\begin{equation}
\mathcal{L}_{\text{ALD}} = \mathcal{L}_{\text{o}}(y; \theta, \sigma, \kappa) + \mathcal{L}_{\text{c}}(y; \theta, \sigma, \kappa) =
- \sum_{n \in \mathcal{D}_{\text{o}}} \log f_{\text{ALD}}(y_n \mid \mathbf{x}_n) 
- \sum_{n \in \mathcal{D}_{\text{c}}} \log \left( 1 - F_{\text{ALD}}(y_n \mid \mathbf{x}_n) \right)
\end{equation}

where $\mathcal{D}_{\text{O}}$ and $\mathcal{D}_{\text{C}}$ are the subsets of $\mathcal{D}$ for which $e=1$ and $e=0$, respectively. 
The first term maximizes the likelihood \(f_{\text{ALD}}(t \mid \mathbf{x})\) for the observed data, while the second term maximizes the survival probability \(S_{\text{ALD}}(t \mid \mathbf{x})\) for the censored data. 
To achieve this, the parameters $\theta, \sigma, \kappa$ predicted by a multi-layer perceptron (MLP) conditioned on the input features, $\mathbf{x}$, enabling the model to adapt flexibly to varying input distributions. 
The observed component \(\mathcal{L}_{\text{o}}(y; \theta, \sigma, \kappa)\) is defined as:

\begin{equation}
\mathcal{L}_{\text{o}}(y; \theta, \sigma, \kappa) = \log \sigma - \log \frac{\kappa}{\kappa^2 + 1} 
+ \frac{\sqrt{2}}{\sigma} 
\begin{cases} 
\kappa (y - \theta), & \text{if } y \geq \theta, \\[10pt]
\frac{1}{\kappa} (\theta - y), & \text{if } y < \theta.
\end{cases}
\end{equation}

The censored loss component \(\mathcal{L}_{\text{c}}(y; \theta, \sigma, \kappa)\) is computed using the survival probability function:

\begin{equation}
\mathcal{L}_{\text{c}}(y; \theta, \sigma, \kappa) = 
\begin{cases} 
\log(\kappa^2 + 1) + \frac{\sqrt{2}}{\sigma} \kappa (y - \theta), & \text{if } y \geq \theta, \\[12pt]
\log(\kappa^2 + 1) - \log\left[1 + \kappa^2\left(1 - \exp\left(-\frac{\sqrt{2}}{\sigma \kappa} (\theta - y)\right)\right)\right], & \text{if } y < \theta.
\end{cases}
\end{equation}

\textbf{Proposition 2. (Mean, Mode, Variance of \( Y \))} The mean, mode, variance of \( Y \) are given by:

\begin{equation}
E[Y] = \theta + \frac{\sigma}{\sqrt{2}} \left( \frac{1}{\kappa} - \kappa \right)
\end{equation}
\begin{equation}
\mathrm{Mode}[Y] = \theta
\end{equation}
\begin{equation}
\mathrm{Var}[Y] = \frac{\sigma^2}{2} \left( \frac{1}{\kappa^2} + \kappa^2 \right)
\end{equation}

\textbf{Proposition 3. (Quantiles of \( Y \))}  
Let \( \theta^{\text{ALD}}_q \) denotes the \( q \)-th quantile of \( Y \). Then, the quantiles can be expressed as:

\begin{equation}
\theta^{\text{ALD}}_q =
\begin{cases} 
\theta + \frac{\sigma \kappa}{\sqrt{2}} \log \left[\frac{1 + \kappa^2}{\kappa^2} q \right], & \text{if } q \in \left(0, \frac{\kappa^2}{1 + \kappa^2}\right], \\[10pt]
\theta - \frac{\sigma}{\sqrt{2}\kappa} \log \left[(1 + \kappa^2)(1 - q) \right], & \text{if } q \in \left( \frac{\kappa^2}{1 + \kappa^2}, 1 \right).
\end{cases}
\end{equation}







\subsection{Analysis of All the Baselines}
\label{appendix:a2}
\textbf{CQRNN.}  
CQRNN \cite{pearce2022censored} combines the likelihood of the Asymmetric Laplace Distribution, \( f_\text{ALD}(t \mid \mathbf{x}) \), with the re-weighting scheme \( w \) introduced by Portnoy \cite{neocleous2006correction}. 
For the observed data, CQRNN employs the Maximum Likelihood Estimation (MLE) approach to directly maximize the likelihood of the Asymmetric Laplace Distribution $\mathcal{AL}(\theta, \sigma, q)$. 
The likelihood is defined over all quantiles of interest. 
For censored data, CQRNN splits each censored data point into two pseudo data points: one at the censoring location $y=c$ and another at a large pseudo value \( y^* \). 
This approach enables the formulation of a weighted likelihood for censored data, resulting in the following loss function:

\begin{equation}
\mathcal{L}_{\text{CQR}} = \mathcal{L}_{\text{o}}(y; \theta, \sigma, q) + \mathcal{L}_{\text{c}}(y, y^*; \theta, \sigma, q, w)
\end{equation}

where \(\mathcal{L}_{\text{o}}\) represents the negative log-likelihood for observed data, and \(\mathcal{L}_{\text{c}}\) accounts for the weighted negative log-likelihood of censored data using the re-weighting scheme. Expanding this, the loss can be expressed as:

\begin{equation}
\mathcal{L}_{\text{CQR}} = 
- \sum_{n \in \mathcal{D}_o} \log f_{\text{ALD}}(y_n \mid \mathbf{x}_n) 
- \sum_{n \in \mathcal{D}_c} [\log f_{\text{ALD}}(y_n \mid \mathbf{x}_n) + (1-w)f_{\text{ALD}}(y^* \mid \mathbf{x}_n)].
\end{equation}

where $\mathcal{D}_{\text{o}}$ and $\mathcal{D}_{\text{c}}$ are the subsets of $\mathcal{D}$ for which $e=1$ and $e=0$, respectively. Here, CQRNN utilizes the Asymmetric Laplace Distribution \(\mathcal{AL}(\theta, \sigma, q)\) to model the data. The Asymmetric Laplace Distribution, denoted as \(\mathcal{AL}(\theta, \sigma, \kappa)\), can be reparameterized as \(\mathcal{AL}(\theta, \sigma, q)\) to facilitate quantile regression within a Bayesian inference framework \cite{yu2001bayesian}, where \(q \in (0, 1)\) is the percentile parameter that represents the desired quantile. The relationship between \(q\) and \(\kappa\) is given by:

\begin{equation}
q = \frac{\kappa^2}{\kappa^2 + 1}.
\end{equation}

Thus, the probability density function for \(Y \sim \mathcal{AL}(\theta, \sigma, q)\) is:

\begin{equation}
f_{\text{ALD}}(y; \theta, \sigma, q) = \frac{q(1 - q)}{\sigma} 
\begin{cases} 
\exp\left(\frac{q}{\sigma}(\theta - y)\right), & \text{if } y \geq \theta, \\[10pt]
\exp\left(\frac{(1 - q)}{\sigma}(y - \theta)\right), & \text{if } y < \theta.
\end{cases}
\end{equation}

And the cumulative distribution function is:

\begin{equation}
F_{\text{ALD}}(y; \theta, \sigma, q) = 
\begin{cases} 
1 - (1 - q)\exp\left(\frac{q}{\sigma}(\theta - y)\right), & \text{if } y \geq \theta, \\[10pt]
q \exp\left(\frac{1 - q}{\sigma}(y - \theta)\right), & \text{if } y < \theta.
\end{cases}
\end{equation}

Thus, the negative log-likelihood $\mathcal{L}_{\text{QR}}(y; \theta, \sigma, q)$ then can be explicitly derived as:

\begin{equation}
\mathcal{L}_{\text{QR}}(y; \theta, \sigma, q) = \log \sigma - \log [q(1-q)] + \frac{1}{\sigma}
\begin{cases} 
q (y - \theta), & \text{if } y \geq \theta \\
(1-q)(\theta - y), & \text{if } y < \theta
\end{cases}
\end{equation}

In their implementation, the scale parameter \( \sigma \) is omitted, and the percentile parameter \( q \) is predefined, typically set to values such as \( q = \{0.1, 0.2, \dots, 0.9\} \). 
A multi-layer perceptron (MLP) in CQRNN, conditioned on the input features \( \mathbf{x} \), predicts \( \theta_q \) for the predefined quantile values, corresponding to the location parameter \( \theta \). 
The negative log-likelihood $\mathcal{L}_{\text{QR}}(y; \theta, \sigma, q)$ is then further simplified as:

\begin{equation}
\mathcal{L}_{\text{QR}}(y; \theta_q, q) =  
\begin{cases} 
q (y - \theta_q), & \text{if } y \geq \theta_q, \\[12pt]
(1 - q)(\theta_q - y), & \text{if } y < \theta_q.
\end{cases}
= (y - \theta_q)(q - \mathbb{I}[\theta_q > y]).
\end{equation}

This formulation is also referred to as the pinball loss or ``checkmark'' loss \cite{koenker1978regression}, which is widely used in quantile regression to directly optimize the $q$-th quantile estimate. For censored data, CQRNN adopts Portnoy’s estimator \cite{neocleous2006correction}, which minimizes a specific objective function tailored for censored quantile regression. This approach introduces a re-weighting scheme to handle all censored data, with the formula defined as:

\begin{equation}
\mathcal{L}_{\text{c}}(y, y^*; \theta_q, q, w) = w \mathcal{L}_{\text{QR}}(y; \theta_q, q) + (1-w)\mathcal{L}_{\text{QR}}(y^*; \theta_q, q),
\end{equation}

where \( y^* \) is a pseudo value set to be significantly larger than all observed values of \( y \) in the dataset. Specifically, it is defined as \( y^* = 1.2 \max_i y_i \) in CQRNN \cite{pearce2022censored}. The weight parameter \( w \) is apportioned between each pair of pseudo-data points as:

\begin{equation}
w = \frac{q - q_c}{1 - q_c},
\end{equation}

where \( q_c \) is the quantile at which the data point was censored (\( e = 0, y = c \)) with respect to the observed value distribution, i.e., \( p(o < c \mid \mathbf{x}) \). 
However, the exact value of \( q_c \) is not accessible in practice. 
To address this issue, CQRNN approximates \( q_c \) using the proportion \( q \) corresponding to the quantile that is closest to the censoring value \( c \), based on the distribution of observed events \( y \), which are readily available.

\textbf{LogNormal MLE.} LogNormal MLE \cite{hoseini2017comparison} enhances parameter estimation using neural networks for LogNormal distributions. Specifically, a random variable \( Y \) follows a LogNormal distribution if the natural logarithm of \( Y \), denoted as \( \ln(Y) \), follows a Normal distribution, {\em i.e.}, \( \ln(Y) \sim \mathcal{N}(\mu, \eta^2) \). Here, \( \mu \) represents the mean, and \( \eta \) is the standard deviation (SD) of the normal distribution. The probability density function of the LogNormal distribution is given by:

\begin{equation}
f_{\text{LogNormal}}(y; \mu, \eta) = \frac{1}{y \eta \sqrt{2\pi}} 
\exp\left(-\frac{(\ln y - \mu)^2}{2\eta^2}\right)
\end{equation}

where \( y > 0 \) and \( \eta > 0 \). The cumulative distribution function is expressed as:

\begin{equation}
F_{\text{LogNormal}}(y, \mu, \eta) = \Phi\left(\frac{\ln(y) - \mu}{\eta}\right)
\end{equation}

where \( \Phi(z) \) is the standard normal cumulative distribution function:

\begin{equation}
\Phi(z) = \frac{1}{\sqrt{2\pi}} \int_{-\infty}^z \exp\left(-\frac{t^2}{2}\right) dt
\end{equation}

The maximum likelihood estimation (MLE) loss with censored data is then defined as:

\begin{equation}
\mathcal{L}_{\text{LogNormal}} = 
- \sum_{n \in \mathcal{D}_o} \log f_{\text{LogNormal}}(y_n \mid \mathbf{x}_n) 
- \sum_{n \in \mathcal{D}_c} \log \left( 1 - F_{\text{LogNormal}}(y_n \mid \mathbf{x}_j) \right).
\end{equation}

A multi-layer perceptron (MLP) in LogNormal MLE, conditioned on the input features \( \mathbf{x} \), is used to predict the mean \( \mu \) and the standard deviation \( \eta \) of the corresponding normal distribution. The quantiles \( \theta^{\text{LogNormal}}_q \) for the LogNormal distribution can be expressed as:

\begin{equation}
\theta^{\text{LogNormal}}_q = \exp(\mu + \eta\Phi^{-1}(q)),
\end{equation}

where \( \Phi^{-1}(q) \) is the inverse CDF (quantile function) of the standard normal distribution.

\textbf{DeepSurv.} DeepSurv \cite{katzman2018deepsurv} is a semi-parametric survival model based on the Cox proportional hazards framework, leveraging deep neural networks for feature representation. A multi-layer perceptron (MLP) in DeepSurv, conditioned on the input features \( \mathbf{x} \), is used to predict the log hazard function \( h(\mathbf{x}) \):

\begin{equation}
\lambda(t \mid \mathbf{x}) = \lambda_0(t) e^{h(\mathbf{x})}
\end{equation}

where \( \lambda_0(t) \) is the baseline hazard function. The hazard function is defined as:

\begin{equation}
\lambda(t \mid \mathbf{x}) = \lim_{\Delta t \to 0} \frac{P(t \leq T < t + \Delta t \mid T \geq t, \mathbf{x})}{\Delta t}
\end{equation}

This can be rewritten as:

\begin{equation}
\lambda(t \mid \mathbf{x}) = -\frac{d S(t \mid \mathbf{x}) / dt}{S(t \mid \mathbf{x})}
\end{equation}

where \( S(t \mid \mathbf{x}) = P(T > t \mid \mathbf{x}) \) is the survival function. By integrating both sides, we have:

\begin{equation}
\int \lambda(t \mid \mathbf{x}) \, dt = \int -\frac{d S(t \mid \mathbf{x})}{S(t \mid \mathbf{x})}
\end{equation}

which simplifies to:

\begin{equation}
\Lambda(t \mid \mathbf{x}) = -\log S(t \mid \mathbf{x}) + C
\end{equation}

where \( C \) is the constant of integration and \( \Lambda(t \mid \mathbf{x}) \) is the cumulative hazard function:

\begin{equation}
\Lambda(t \mid \mathbf{x}) = \Lambda_0(t) e^{h(\mathbf{x})}
\end{equation}

where \( \Lambda_0(t) \) is the baseline cumulative hazard function. For survival analysis, \( C \) is typically set to 0 when starting from \( t = 0 \). Thus, the survival function can be expressed as:

\begin{equation}
S(t \mid \mathbf{x}) = e^{-\Lambda(t \mid \mathbf{x})} = e^{-\Lambda_0(t) e^{h(\mathbf{x})}} = \left[S_0(t)\right]^{e^{h(\mathbf{x})}}
\end{equation}

where \( S_0(t) \) is the baseline survival function, typically estimated by the Kaplan-Meier method \cite{kaplan1958nonparametric} using the training data. The cumulative distribution function (CDF) can then be derived as:

\begin{equation}
F_{\text{DeepSurv}}(t \mid \mathbf{x}) = 1 - S(t \mid \mathbf{x}) = 1 - \left[S_0(t)\right]^{e^{h(\mathbf{x})}}
\end{equation}

The quantiles \( \theta^{\text{DeepSurv}}_q \) for DeepSurv can be obtained from the inverse CDF $F_{\text{DeepSurv}}^{-1}(t \mid \mathbf{x})$ (quantile function).

\textbf{DeepHit.} A multi-layer perceptron (MLP) in DeepHit \cite{lee2018deephit}, conditioned on the input features \( \mathbf{x} \), is used to predict the probability distribution \( f(t \mid \mathbf{x}) \) over event times using a fully non-parametric approach. The quantiles \( \theta^{\text{DeepHit}}_q \) can be obtained from the inverse cumulative distribution function \( F_{\text{DeepHit}}^{-1}(t \mid \mathbf{x}) \), where \( F_{\text{DeepHit}}(t \mid \mathbf{x}) = \sum f_{\text{DeepHit}}(t \mid \mathbf{x}) \).

\section{Experimental Details}
This section provides additional details about the experiments conducted. The experiments were implemented using the \href{https://pytorch.org/}{PyTorch framework}. Detailed information about the datasets, metrics, baselines and implementation details can be found in Appendix \ref{appendix:b1}, Appendix \ref{appendix:b2}, and Appendix \ref{appendix:b3}, respectively.

\textbf{Hardware.} All experiments were conducted on a MacBook Pro with an Apple M3 Pro chip, featuring 12 cores (6 performance and 6 efficiency cores) and 18 GB of memory. CPU-based computations were utilized for all experiments, as the models primarily relied on fully-connected neural networks.

\subsection{Datasets}
\label{appendix:b1}
Our datasets are designed following the settings outlined in \citet{pearce2022censored}. The first type of dataset consists of synthetic target data with synthetic censoring. In these datasets, the input features, $\mathbf{x}$, are generated uniformly as $\mathbf{x} \sim \mathcal{U} (0, 2)^D$, where $D$ denotes the number of features. The observed variable, $o \sim p(o \mid \mathbf{x})$, and the censored variable, $c \sim p(c \mid \mathbf{x})$, follow distinct distributions, with their parameters varying based on the specific dataset configuration. Table \ref{tab:synthetic dataset} provides detailed descriptions of the distributions for the observed and censored variables. Additionally, the coefficient vector used in some datasets is defined as $\beta = [0.8, 0.6, 0.4, 0.5, -0.3, 0.2, 0.0, -0.7]$.

\begin{table}[ht]
\caption{Characteristics of synthetic datasets encompassing the number of features, parameterized distributions of observed variables, and censored variables, as utilized in the experimental framework.}
\label{tab:synthetic dataset}
\centering
\resizebox{1\textwidth}{!}{
\begin{tabular}{ccccc}
\toprule
\textbf{Synthetic Dataset} & \textbf{Feats ($D$)} & \textbf{Observed Variables $o \sim p(o \mid \mathbf{x})$} & \textbf{Censored Variables $c \sim p(c \mid \mathbf{x})$}\\
\midrule
Norm linear       & 1 & $\mathcal{N}(2\mathbf{x}+10, (\mathbf{x}+1)^2)$ & $\mathcal{N}(4\mathbf{x}+10, (0.8\mathbf{x}+0.4)^2)$    \\
Norm non-linear   & 1 & $\mathcal{N}(\mathbf{x}\mathrm{sin}(2\mathbf{x})+10, (0.5\mathbf{x}+0.5)^2)$ & $\mathcal{N}(2\mathbf{x}+10, 2^2)$ \\
Exponential       & 1 & $\mathrm{Exp}(2\mathbf{x}+4)$ & $\mathrm{Exp}(-3\mathbf{x}+15)$  \\
Weibull           & 1 & $\mathrm{Weibull}(\mathbf{x}\mathrm{sin}(2\mathbf{x}-2)+10, 5)$ & $\mathrm{Weibull}(-3\mathbf{x}+20, 5)$  \\
LogNorm           & 1 & $\mathrm{LogNorm}((\mathbf{x}-1)^2, \mathbf{x}^2)$  & $\mathcal{U} (0, 10)$ \\
Norm uniform      & 1 & $\mathcal{N}(2\mathbf{x}\mathrm{cos}(2\mathbf{x})+13, (\mathbf{x}+0.5)^2)$ & $\mathcal{U} (0, 18)$ \\
Norm heavy        & 4 & $\mathcal{N}(3\mathbf{x}_0+\mathbf{x}_1^2-\mathbf{x}_2^2+2\mathrm{sin}(\mathbf{x}_2\mathbf{x}_3)+6, (\mathbf{x}+0.5)^2)$ & $\mathcal{U} (0, 12)$  \\
Norm med          & 4 & ---"--- & $\mathcal{U} (0, 20)$  \\
Norm light        & 4 & ---"--- & $\mathcal{U} (0, 40)$  \\
Norm same         & 4 & ---"--- & Equal to observed dist. \\
LogNorm heavy     & 8 & $\mathrm{LogNorm}(\sum_{i=1}^{8} \beta_i\mathbf{x}_i, 1)/10$ & $\mathcal{U} (0, 0.4)$  \\
LogNorm med       & 8 & ---"--- & $\mathcal{U} (0, 1.0)$  \\
LogNorm light     & 8 & ---"--- & $\mathcal{U} (0, 3.5)$   \\
LogNorm same      & 8 & ---"--- & Equal to observed dist.  \\
\bottomrule
\end{tabular}
}
\end{table}

The other type of dataset comprises real-world target data with real censoring, sourced from various domains and characterized by distinct features, sample sizes, and censoring proportions:
\begin{itemize}
    \item \textbf{METABRIC (Molecular Taxonomy of Breast Cancer International Consortium):} This dataset contains genomic and clinical data for breast cancer patients. It includes 9 features, 1523 training samples, and 381 testing samples, with a censoring proportion of 0.42. Retrieved from \href{https://github.com/jaredleekatzman/DeepSurv/}{the DeepSurv Repository}.

    \item \textbf{WHAS (Worcester Heart Attack Study):} This dataset focuses on predicting survival following acute myocardial infarction. It includes 6 features, 1310 training samples, and 328 testing samples, with a censoring proportion of 0.57. Retrieved from \href{https://github.com/jaredleekatzman/DeepSurv/}{the DeepSurv Repository}.

    \item \textbf{SUPPORT (Study to Understand Prognoses Preferences Outcomes and Risks of Treatment):} This dataset provides survival data for critically ill hospitalized patients. It includes 14 features, 7098 training samples, and 1775 testing samples, with a censoring proportion of 0.32. Covariates include demographic information and basic diagnostic data. Retrieved from \href{https://github.com/jaredleekatzman/DeepSurv/}{the DeepSurv Repository}.

    \item \textbf{GBSG (Rotterdam \& German Breast Cancer Study Group):} Originating from the German Breast Cancer Study Group, this dataset tracks survival outcomes of breast cancer patients. It includes 7 features, 1785 training samples, and 447 testing samples, with a censoring proportion of 0.42. Retrieved from \href{https://github.com/jaredleekatzman/DeepSurv/}{the DeepSurv Repository}. 

    \item \textbf{TMBImmuno (Tumor Mutational Burden and Immunotherapy):} This dataset predicts survival time for patients with various cancer types using clinical data. It includes 3 features, 1328 training samples, and 332 testing samples, with a censoring proportion of 0.49. Covariates include age, sex, and mutation count. Retrieved from \href{https://www.cbioportal.org/study/clinicalData?id=tmb_mskcc_2018}{the cBioPortal}.

    \item \textbf{BreastMSK:} Derived from the Memorial Sloan Kettering Cancer Center, this dataset focuses on predicting survival time for breast cancer patients using tumor-related information. It includes 5 features, 1467 training samples, and 367 testing samples, with a censoring proportion of 0.77.  Retrieved from \href{https://www.cbioportal.org/study/clinicalData?id=breast_msk_2018}{the cBioPortal}.

    \item \textbf{LGGGBM:} This dataset integrates survival data from low-grade glioma (LGG) and glioblastoma multiforme (GBM), frequently used for model validation in cancer genomics. It includes 5 features, 510 training samples, and 128 testing samples, with a censoring proportion of 0.60. Retrieved from \href{https://www.cbioportal.org/study/clinicalData?id=lgggbm_tcga_pub}{the cBioPortal}.
\end{itemize}

\subsection{Metrics}
\label{appendix:b2}
We employ nine distinct evaluation metrics to assess model performance comprehensively: Mean Absolute Error (MAE), Integrated Brier Score (IBS) \cite{graf1999assessment}, Harrell’s C-Index \cite{harrell1982evaluating}, Uno’s C-Index \cite{uno2011c}, censored D-calibration (CensDcal) \cite{haider2020effective}, along with the slope and intercept derived from two versions of censored D-calibration (Cal $[S(t|\mathbf{x})]$ (Slope), Cal$[S(t|\mathbf{x})]$(Intercept), Cal$[f(t|\mathbf{x})]$(Slope), and Cal$[f(t|\mathbf{x})]$(Intercept)). These metrics provide a holistic evaluation framework, effectively capturing the survival models' predictive accuracy, discriminative ability, and calibration quality.

\begin{itemize}
    \item \textbf{MAE:} 
    \begin{equation} 
        \mathrm{MAE} = \frac{1}{N} \sum_{i=1}^{N}|y_i - \tilde{y}_i| 
    \end{equation}
    where $y_i$ represents the observed survival times, $\tilde{y}_i$ denotes the predicted survival times, and $N$ is the total number of data points in the test set.

    \item \textbf{IBS:} 
    \begin{equation} 
        \mathrm{BS}(t) = \frac{1}{N} \sum_{i=1}^N \left[ \frac{\left(1 - \tilde{F}(t \mid \mathbf{x}_i)\right)^2 \mathbb{I}\left(y_i\leq t, e_i = 1\right)}{\tilde{G}(y_i)} + \frac{\tilde{F}(t \mid \mathbf{x}_i)^2 \mathbb{I}\left(y_i> t\right)}{\tilde{G}(t)} \right]
    \end{equation}
    \begin{equation}    
        \mathrm{IBS} = \frac{1}{t_2 - t_1} \int_{t_1}^{t_2} \mathrm{BS}(y) \, dy
    \end{equation}
    where $\mathrm{BS}(t)$ represents the Brier score at time $t$, and 100 time points are evenly selected from the 0.1 to 0.9 quantiles of the $y$-distribution in the training set. $\tilde{F}(t \mid \mathbf{x}_i)$ denotes the estimated cumulative distribution function of the survival time for test subjects, $\mathbb{I}(\cdot)$ is the indicator function, and $e_i$ is the event indicator ($e_i = 1$ if the event is observed). $\mathbf{x}_i$ represents the covariates, and $\tilde{G}(\cdot)$ refers to the Kaplan-Meier estimate \cite{kaplan1958nonparametric} of the censoring survival function.
    \item \textbf{Harrell’s C-Index:} 
    \begin{equation}
        \mathrm{C_H} = P(\phi_i > \phi_j \mid y_i < y_j, e_i = 1) = \frac{\sum_{i \neq j} \big[\mathbb{I}(\phi_i > \phi_j) + 0.5 \ast \mathbb{I}(\phi_i = \phi_j)\big] \mathbb{I}(y_i < y_j) e_i} {\sum_{i \neq j} \mathbb{I}(y_i < y_j) e_i}
    \end{equation}
    where $\phi_i = \tilde{S}(y_i \mid \mathbf{x}_i) = 1 - \tilde{F}(t \mid \mathbf{x}_i)$ represents the risk score predicted by the survival model. For implementation, we utilize the \texttt{concordance\_index\_censored} function from the \texttt{sksurv.metrics} module, as documented in the \href{https://scikit-survival.readthedocs.io/en/stable/api/generated/sksurv.metrics.concordance_index_censored.html}{scikit-survival API}.
    \item \textbf{Uno’s C-Index:} 
    \begin{align}
        \mathrm{C_U} &= P(\phi_i > \phi_j \mid y_i < y_j, y_i < y_\tau) \nonumber \\
        &= 
        \frac{\sum_{i=1}^n \sum_{j=1}^n G(y_i)^{-2} [\mathbb{I}(\phi_i > \phi_j) + 0.5 \ast \mathbb{I}(\phi_i = \phi_j)] \mathbb{I}(y_i < y_j, y_i < y_\tau) e_i}
        {\sum_{i=1}^n \sum_{j=1}^n G(y_i)^{-2} \mathbb{I}(y_i < y_j, y_i < y_\tau) e_i}.
    \end{align}
    where $y_\tau$ is the cutoff value for the survival time. For implementation, we utilize the \texttt{concordance\_index\_ipcw} function from the \texttt{sksurv.metrics} module, as documented in the \href{https://scikit-survival.readthedocs.io/en/stable/api/generated/sksurv.metrics.concordance_index_ipcw.html}{scikit-survival API}.
    \item \textbf{CensDcal:} 
    \begin{equation}
        \mathrm{CensDcal} = 100 \times \sum_{j=1}^{10} \left( (q_{j+1} - q_j) - \frac{1}{N} \zeta \right)^2,
    \end{equation}
    where $\zeta$ is defined by \cite{goldstein2020x} as:
    \begin{equation}
        \zeta = \sum_{i \in \mathcal{S}_{\text{observed}}} \mathbb{I}[\tilde{\theta}_{i, q_j} < y_i \leq \tilde{\theta}_{i,q_{j+1}}] 
        + \sum_{i \in \mathcal{S}_{\text{censored}}} 
        \frac{(q_{j+1} - q_i) \mathbb{I}[\tilde{\theta}_{i, q_j} < y_i \leq \tilde{\theta}_{i,q_{j+1}}]}{1 - q_i} 
        + \frac{(q_{j+1} - q_j) \mathbb{I}[q_i < q_j]}{1 - q_i}.
    \end{equation}
    Here, the percentile parameter $q_j$ is predefined as $[0.1, 0.2, \ldots, 0.9]$ at the outset, and $q_i$ is the quantile at which the data point was censored ($e=0, y=c$) with respect to the observed value distribution, {\em i.e.}, $p(o < c \mid \mathbf{x})$. $\tilde{\theta}_{i, q_j}$ represents the estimated $q$th quantile of $y_i$.
    \item \textbf{Slope \& Intercept:} The Slope and Intercept metrics evaluate the calibration quality of predicted survival quantiles relative to observed data under censoring. We utilize the \texttt{np.polyfit} function from the \texttt{NumPy} module, as documented in the \href{https://numpy.org/doc/stable/reference/generated/numpy.polyfit.html}{NumPy API}, to fit the 10 points $\left\{\left(0.1j, \sum_j \frac{1}{N}\zeta_j\right)\right\}_{j=1}^{10}$ and subsequently obtain the Slope and Intercept metrics.  
    Two versions of the Slope and Intercept (Cal$[S(t|\mathbf{x})]$(Slope), Cal$[S(t|\mathbf{x})]$(Intercept), Cal$[f(t|\mathbf{x})]$(Slope), and Cal$[f(t|\mathbf{x})]$(Intercept)) are calculated, differing in how the quantile intervals are defined:
    \begin{itemize}
        \item \textbf{Version 1 (Measuring $S(t \mid \mathbf{x})$):} The predicted survival probabilities are divided into intervals based on the target proportions, {\em i.e.}, $q = [0.1, 0.2, \ldots, 0.9, 1.0]$. For each quantile interval, the proportion of ground truth values (observed survival times) that fall within the corresponding predicted quantile $\frac{1}{N} \zeta$ is calculated. For example, the ratio for 0.1 ($j=1$) is calculated within the interval $[0, 0.1]$, and for 0.2 ($j=2$), within $[0, 0.2]$. Thus, the horizontal axis represents the target proportions $0.1j$, while the vertical axis represents the observed proportions $\sum_j \frac{1}{N}\zeta_j$ derived from predictions. In the end, this metric is suitable for evaluating the Survival Function $S(t \mid \mathbf{x})$ (or CDF $F(t \mid \mathbf{x})$).
        \item \textbf{Version 2 (Measuring $f(t \mid \mathbf{x})$):} Narrower intervals centered around target proportions are used, {\em i.e.}, $q = [\ldots, 0.4, 0.45, 0.55, 0.6, \ldots]$. For each quantile, the observed proportions are calculated within these narrower intervals. For example, the ratio for 0.1 is calculated within the interval $[0.45, 0.55]$, and for 0.2, within $[0.4, 0.6]$. In the end, this metric is ideal for assessing the probability density function (PDF) $f(t \mid \mathbf{x})$.
    \end{itemize}
\end{itemize}

\subsection{Implementation Details}
\label{appendix:b3}
\textbf{Baselines.} We compare our method against four baselines to evaluate performance and effectiveness: \textbf{LogNorm} \cite{royston2001lognormal}, \textbf{DeepSurv} \cite{katzman2018deepsurv}, \textbf{DeepHit} \cite{lee2018deephit}, and \textbf{CQRNN} \cite{pearce2022censored}. All methods were trained using the same optimization procedure and neural network architecture to ensure a fair comparison. The implementations for \textbf{CQRNN} and \textbf{LogNorm} were sourced from the official CQRNN repository (\href{https://github.com/TeaPearce/Censored_Quantile_Regression_NN}{GitHub Link}). The implementations for \textbf{DeepSurv} and \textbf{DeepHit} were based on the \texttt{pycox.methods} module (\href{https://github.com/havakv/pycox/}{GitHub Link}). 

\textbf{Hyperparameter settings.} All experiments were repeated across 10 random seeds to ensure robust and reliable results. The hyperparameter settings were as follows:
\begin{itemize}
    \item \textbf{Default Neural Network Architecture:} Fully-connected network with two hidden layers, each consisting of 100 hidden nodes, using ReLU activations.
    \item \textbf{Default Epochs:} 200
    \item \textbf{Default Batch Size:} 128
    \item \textbf{Default Learning Rate:} 0.01
    \item \textbf{Dropout Rate:} 0.1
    \item \textbf{Optimizer:} Adam
    \item \textbf{Batch Norm:} FALSE
\end{itemize}

\textbf{Our Method.} We incorporate a residual connection between the shared feature extraction layer and the first hidden layer to enhance gradient flow. To satisfy the parameter constraints of the Asymmetric Laplace Distribution (ALD), the final output layer applies an exponential ($\mathrm{Exp}$) activation function, ensuring that the outputs of the $\theta$, $\sigma$ and $\kappa$ branches remain positive. Each of the two hidden layers contains 32 hidden nodes. A validation set is created by splitting 20\% of the training set. Early stopping is utilized to terminate training when the validation performance ceases to improve.

\textbf{CQRNN.} We followed the hyperparameter settings tuned in the original paper \cite{pearce2022censored}, where three random splits were used for validation (ensuring no overlap with the random seeds used in the final test runs). The following settings were applied:
\begin{itemize}
    \item \textbf{Weight Decay:} 0.0001
    \item \textbf{Grid Size:} 10
    \item \textbf{Pseudo Value:} \( y^* = 1.2 \times \max_i y_i \)
    \item \textbf{Dropout Rate:} 0.333
\end{itemize}
The number of epochs and dropout usage were adjusted based on the dataset type:
\begin{itemize}
    \item \textbf{Synthetic Datasets:}
    \begin{itemize}
        \item \textbf{Norm linear, Norm non-linear, Exponential, Weibull, LogNorm, Norm uniform:} 100 epochs with dropout disabled.
        \item \textbf{Norm heavy, Norm medium, Norm light, Norm same:} 20 epochs with dropout disabled.
        \item \textbf{LogNorm heavy, LogNorm medium, LogNorm light, LogNorm same:} 10 epochs with dropout disabled.
    \end{itemize}
    \item \textbf{Real-World Datasets:}
    \begin{itemize}
        \item \textbf{METABRIC:} 20 epochs with dropout disabled.
        \item \textbf{WHAS:} 100 epochs with dropout disabled.
        \item \textbf{SUPPORT:} 10 epochs with dropout disabled.
        \item \textbf{GBSG:} 20 epochs with dropout enabled.
        \item \textbf{TMBImmuno:} 50 epochs with dropout disabled.
        \item \textbf{BreastMSK:} 100 epochs with dropout disabled.
        \item \textbf{LGGGBM:} 50 epochs with dropout enabled.
    \end{itemize}
\end{itemize}

\textbf{LogNorm.} The output dimensions of the default neural network architecture are 2, where the two outputs represent the mean and standard deviation of a Log-Normal distribution. To ensure the standard deviation prediction is always positive and differentiable, the output representing the standard deviation is passed through a \texttt{SoftPlus} activation function. We followed the hyperparameter settings tuned in the original paper \cite{pearce2022censored}, with a \textbf{Dropout Rate} of 0.333. The number of epochs and dropout usage were adjusted based on the dataset type as follows:

\begin{itemize}
    \item \textbf{Synthetic Datasets:} The same settings as described above for \textbf{CQRNN}.
    \item \textbf{Real-World Datasets:}
    \begin{itemize}
        \item \textbf{METABRIC:} 10 epochs with dropout disabled.
        \item \textbf{WHAS:} 50 epochs with dropout disabled.
        \item \textbf{SUPPORT:} 20 epochs with dropout disabled.
        \item \textbf{GBSG:} 10 epochs with dropout enabled.
        \item \textbf{TMBImmuno:} 50 epochs with dropout disabled.
        \item \textbf{BreastMSK:} 50 epochs with dropout disabled.
        \item \textbf{LGGGBM:} 20 epochs with dropout enabled.
    \end{itemize}
\end{itemize}

\textbf{DeepSurv.} We adhered to the official hyperparameter settings from the \texttt{pycox.methods} module (\href{https://github.com/havakv/pycox/}{GitHub Link}). Each of the two hidden layers contains 32 hidden nodes. A validation set was created by splitting 20\% of the training set. Early stopping was employed to terminate training when the validation performance ceased to improve. Batch normalization was applied.

\textbf{DeepHit.} We adhered to the official hyperparameter settings from the \texttt{pycox.methods} module (\href{https://github.com/havakv/pycox/}{GitHub Link}). Each of the two hidden layers contains 32 hidden nodes. A validation set was created by splitting 20\% of the training set. Early stopping was employed to terminate training when the validation performance ceased to improve. Batch normalization was applied, with additional settings: \texttt{num\_durations} = 100, \texttt{alpha} = 0.2, and \texttt{sigma} = 0.1.

\section{Additional Results} 
\label{appendix:c}
\begin{figure*}[ht]
    \subfigure{
        \includegraphics[width=1\linewidth]{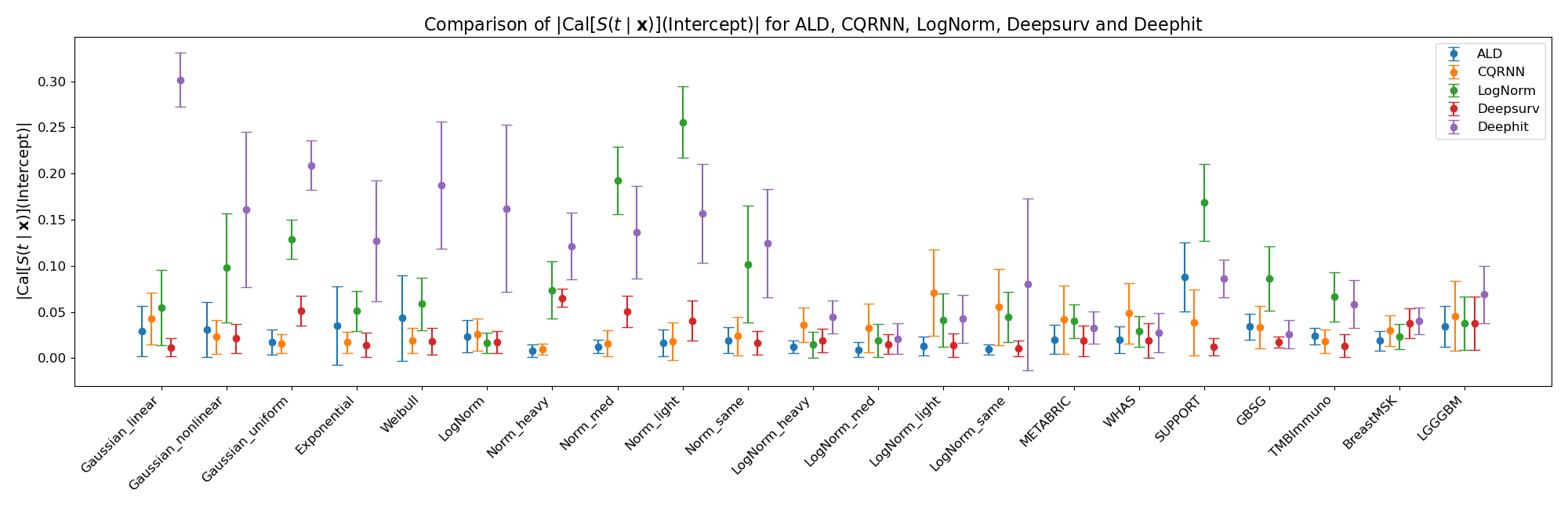}
        \label{fig:Slope2}
    }
    \subfigure{
        \includegraphics[width=1\linewidth]{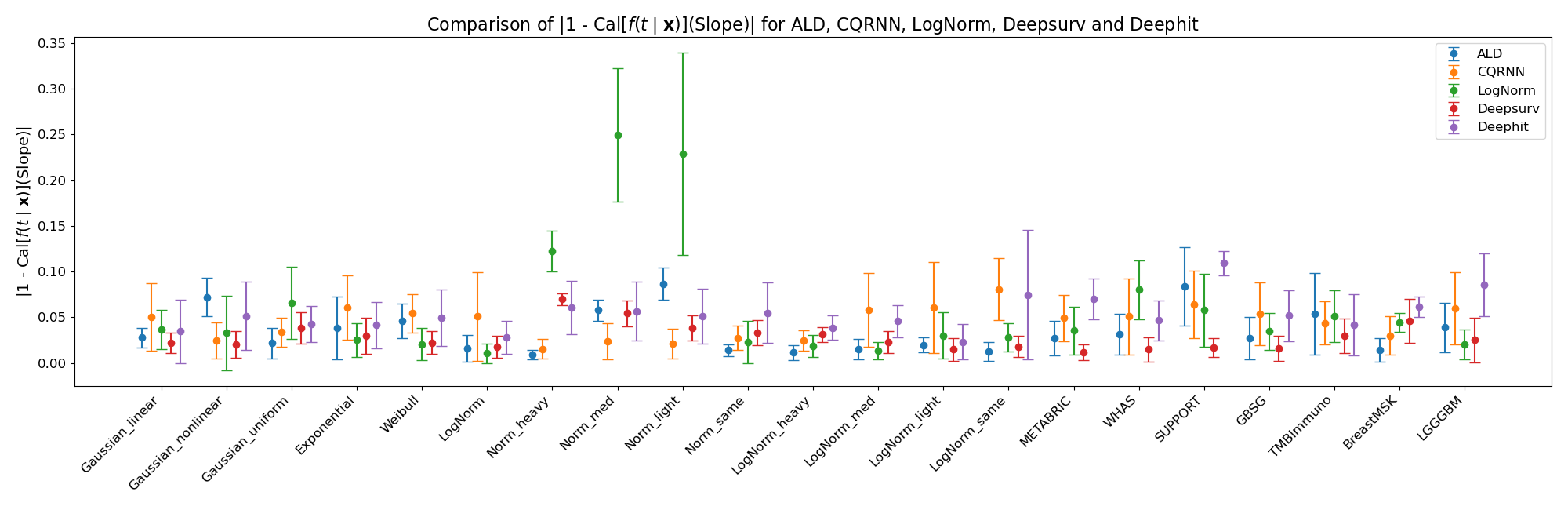}
        \label{fig:Intercept1}
    }
    \subfigure{
        \includegraphics[width=1\linewidth]{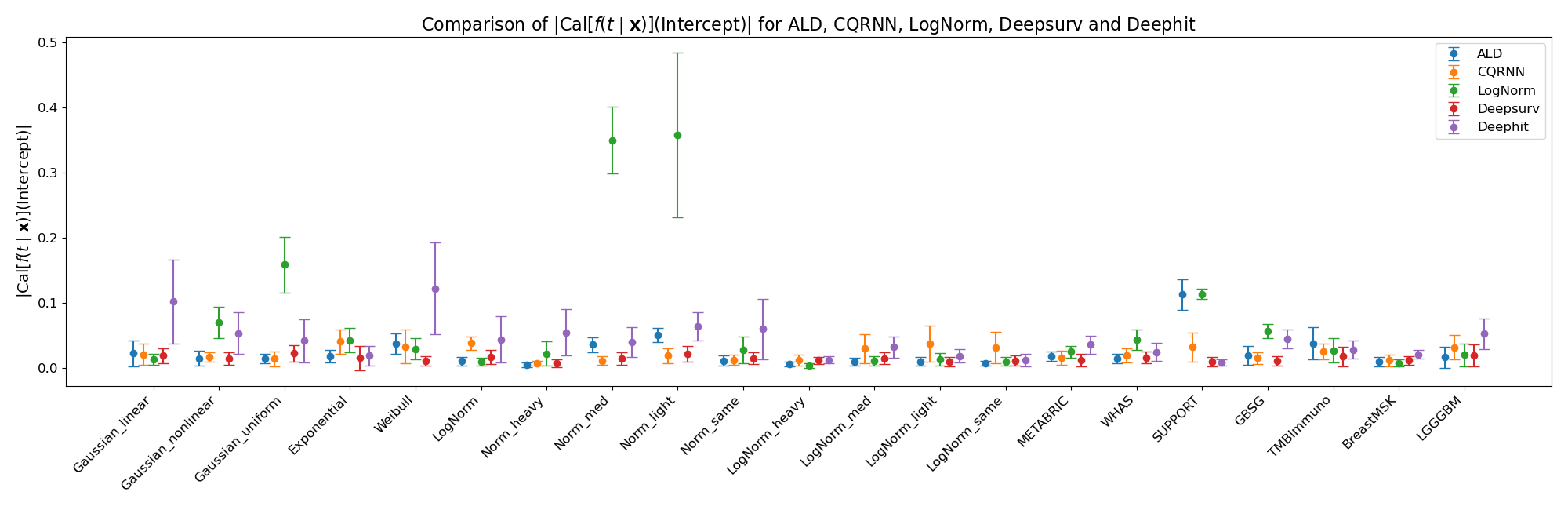}
        \label{fig:Intercept2}
    }
    \caption{Performance on calibration metrics.}
    \label{fig:overall_results}
\end{figure*}

\begin{figure*}[ht] 
    \centering
    \includegraphics[width=\linewidth]{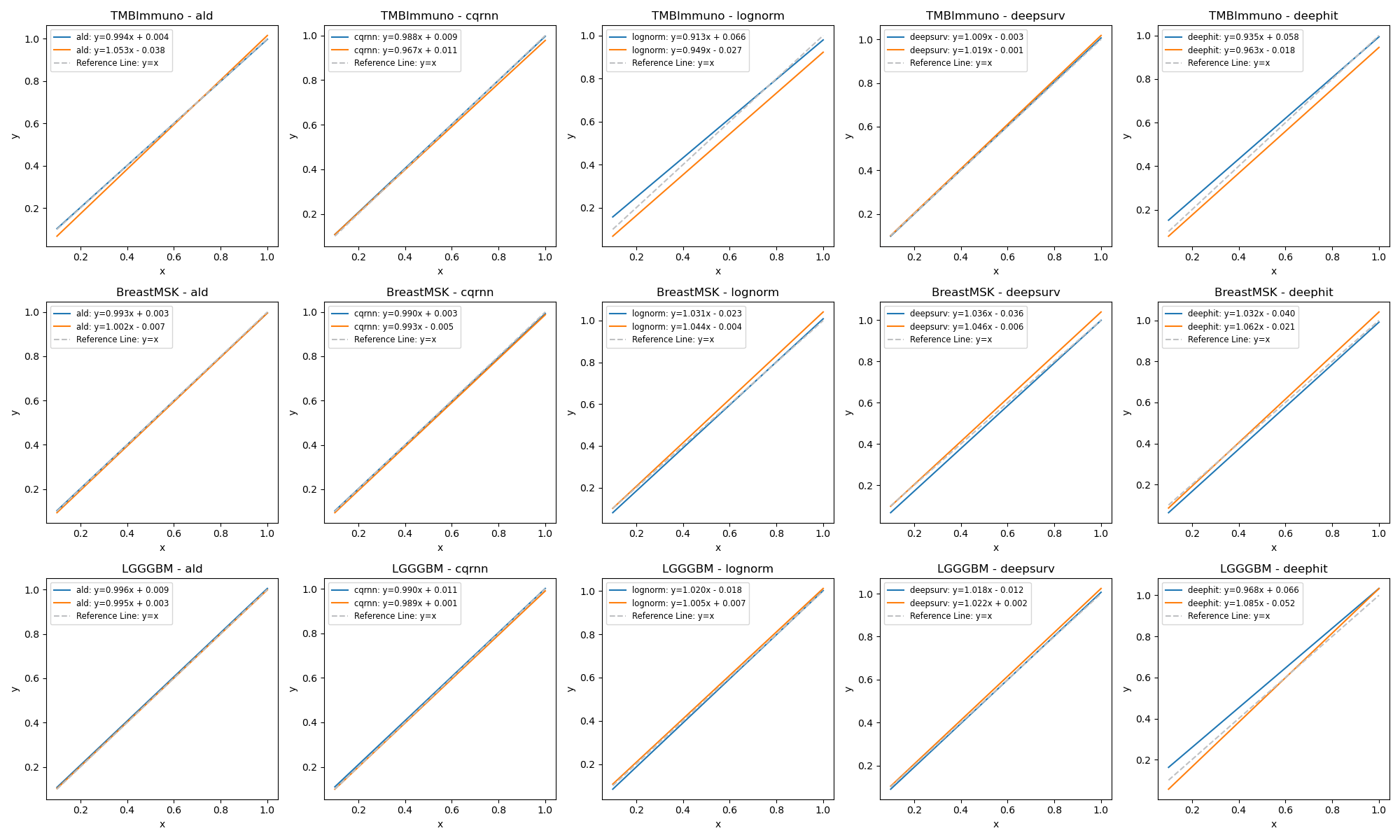} 
    \caption{Calibration Linear Fit. The blue and orange lines represent the curves for Cal$[S(t \mid \mathbf{x})]$ and Cal$[f(t \mid \mathbf{x})]$, respectively. The gray dashed line represents the idealized result where the slope is one and the intercept is zero.}
    \label{fig:lineplot_all4} 
\end{figure*}

\begin{figure*}[ht] 
    \centering
    \includegraphics[width=\linewidth]{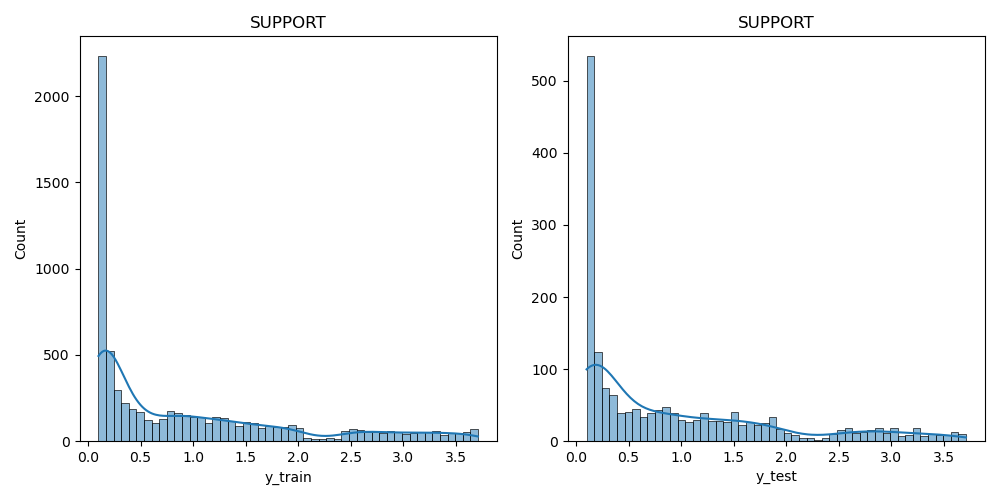} 
    \caption{Distribution of the SUPPORT dataset for the training set and test set.}
    \label{fig:SUPPORT} 
\end{figure*}

This section presents additional results to provide a comprehensive evaluation.
Figure \ref{fig:overall_results} plots 9 distinct evaluation metrics, each presented with error bars for clarity, based on 10 model runs.
Figure \ref{fig:lineplot_all4} illustrates the full results for the calibration linear fit.
Figure \ref{fig:SUPPORT} shows the statistical distribution of the SUPPORT dataset for both the training set and the test set. 
The histograms illustrate the count of observations across different $y$-values, with the blue lines representing density estimates.
Table \ref{tab:overall results} provides the complete results for all datasets, methods, and metrics.
Table \ref{tab:overall results3} summarizes the results for all datasets, focusing on the ALD method (Mean, Median, Mode) across various metrics.
Finally, Table \ref{tab:ald_cdf} presents the 50th, 75th, and 95th percentiles of the CDF estimation for $t=0$, $F_{\text{ALD}}(0 \mid \mathbf{x})$, under the Asymmetric Laplace Distribution.

\textbf{Overall Results.} 
Table \ref{tab:overall results3} summarizes the full results across 21 datasets, comparing our method with 4 baselines across 9 metrics and Figure \ref{fig:overall_results} visualizes these results for a more intuitive comparison. 
In Table \ref{tab:overall results3}, the best performance is highlighted in bold. 
Figure \ref{fig:overall_results} provides a graphical representation of nine distinct evaluation metrics to comprehensively assess predictive performance, including Mean Absolute Error (MAE), Integrated Brier Score (IBS), Harrell’s C-Index, Uno’s C-Index, Censored D-calibration (CensDcal), and the slope and intercept derived from two versions of censored D-calibration (Cal$[S(t|\mathbf{x})]$(Slope), Cal$[S(t|\mathbf{x})]$(Intercept), Cal$[f(t|\mathbf{x})]$(Slope), and Cal$[f(t|\mathbf{x})]$(Intercept)). 
Specifically, the following transformations were applied to enhance the clarity of the results:

\begin{itemize}
    \item MAE and CensDcal were log-transformed to better illustrate their value distributions and differences.
    \item For Cal$[S(t|\mathbf{x})]$(Slope) and Cal$[f(t|\mathbf{x})]$(Slope), \(|1 - \text{Cal}[S(t|\mathbf{x})](\text{Slope})|\) and \(|1 - \text{Cal}[f(t|\mathbf{x})](\text{Slope})|\) were computed to measure their deviation from the ideal value of 1.
    \item For Cal$[S(t|\mathbf{x})]$(Intercept) and Cal$[f(t|\mathbf{x})]$(Intercept), \(|\text{Cal}[S(t|\mathbf{x})](\text{Intercept})|\) and \(|\text{Cal}[f(t|\mathbf{x})](\text{Intercept})|\) were computed to measure their deviation from the ideal value of 0.
\end{itemize}

These transformations allow for a more intuitive comparison of the performance differences across metrics and models. 
In the end, each subfigure in Figure \ref{fig:overall_results} provides a comparison of its corresponding metric. 
The x-axis lists all the datasets, both synthetic datasets ({\em e.g.}, Gaussian linear, exponential, {\em etc}.) and real-world datasets ({\em e.g.}, METABRIC, WHAS, {\em etc}.), while the y-axis indicates the range of corresponding its metric, with error bars representing standard deviation across 10 model runs.

\textbf{Calibration.} 
Figure \ref{fig:lineplot_all4} illustrates the full results of the calibration linear fit, providing a more intuitive and graphical perspective of the calibration performance. 
The horizontal axis represents the target proportions \([0.1, 0.2, \ldots, 0.9, 1.0]\), while the vertical axis denotes the observed proportions derived from the model predictions.

\begin{figure*}[h]
    \centering
    \subfigure{
        \includegraphics[width=1\linewidth]{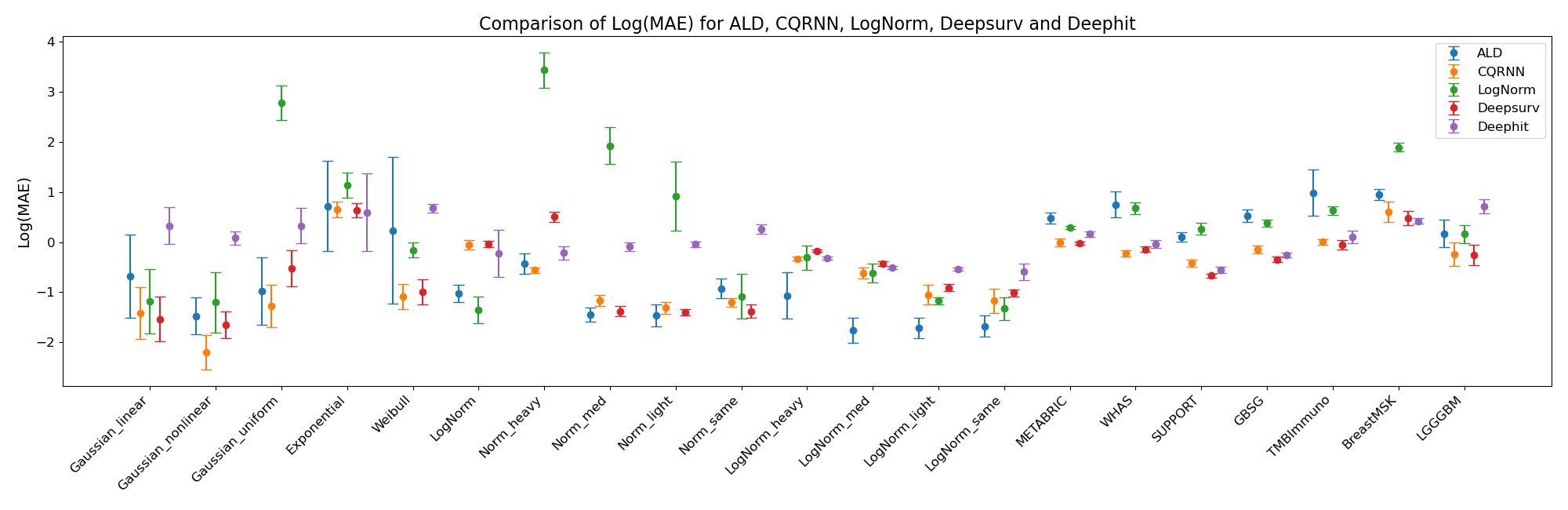}
        \label{fig:MAE}
    }
    \subfigure{
        \includegraphics[width=1\linewidth]{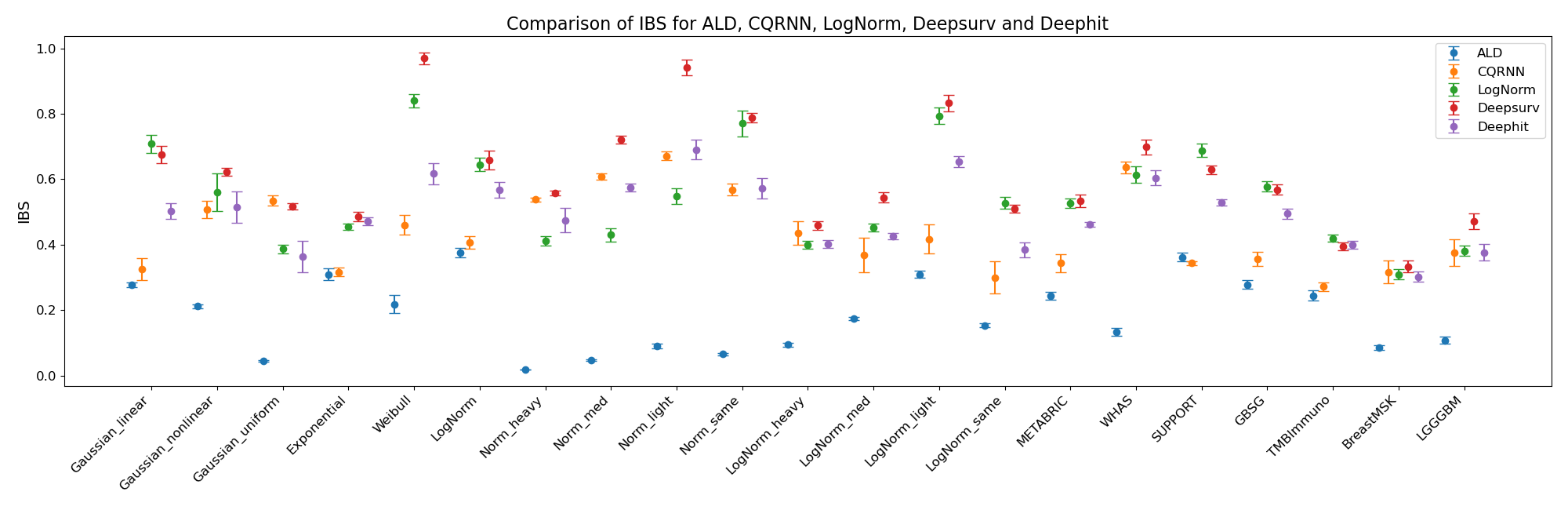}
        \label{fig:IBS}
    }
    \subfigure{
        \includegraphics[width=1\linewidth]{fig/Harrells_C-Index_errorbars.png}
        \label{fig:Harrell’s C-Index}
    }
\end{figure*}

\begin{figure*}[ht]
    \centering 
    \subfigure{
        \includegraphics[width=1\linewidth]{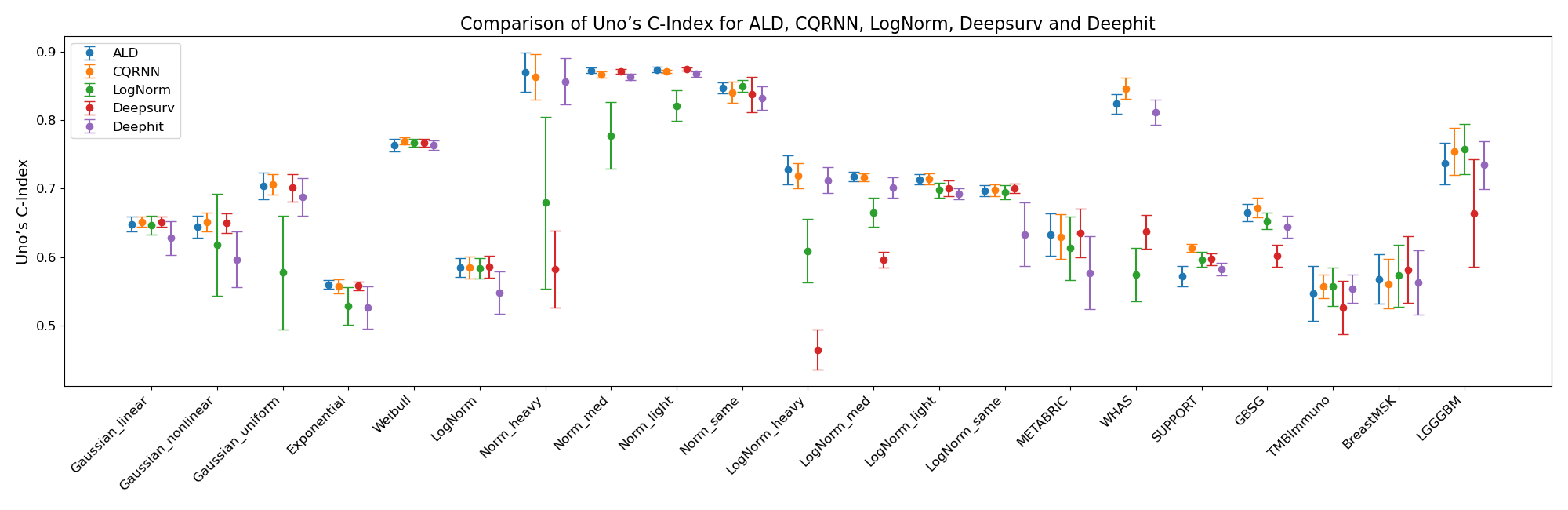}
        \label{fig:Uno’s C-Index}
    }
    \subfigure{
        \includegraphics[width=1\linewidth]{fig/CensDcal_errorbars.png}
        \label{fig:CensDcal}
    }
    \subfigure{
        \includegraphics[width=1\linewidth]{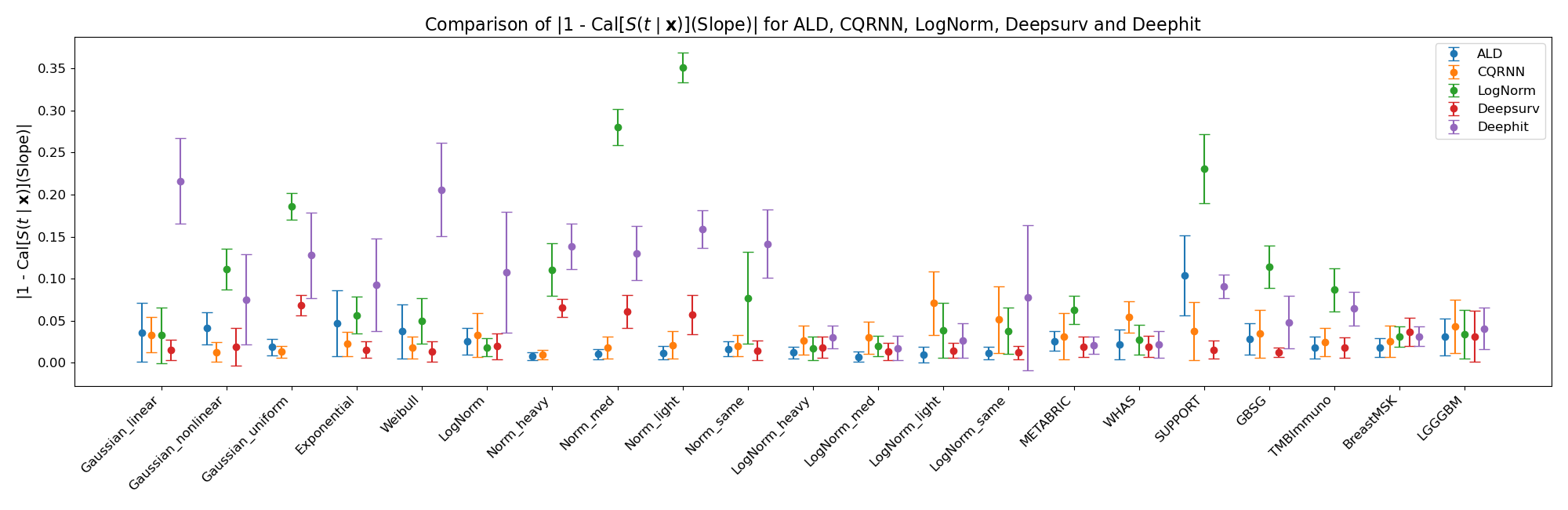}
        \label{fig:Slope1}
    }    
\end{figure*}

\begin{table}[ht]
\caption{Full results table for all datasets, methods, and metrics. The values represent the mean ± 1 standard error for the test set over 10 runs.}
\label{tab:overall results}
\centering
\resizebox{1\textwidth}{!}{
\begin{tabular}{ccccccccccc}
\toprule
\textbf{Dataset} & \textbf{Method} & \textbf{MAE} & \textbf{IBS} & \textbf{Harrell’s C-index} & \textbf{Uno’s C-index} & \textbf{CensDcal} & \textbf{Cal$[S(t|\mathbf{x})]$(Slope)} & \textbf{Cal$[S(t|\mathbf{x})]$(Intercept)} & \textbf{Cal$[f(t|\mathbf{x})]$(Slope)} & \textbf{Cal$[f(t|\mathbf{x})]$(Intercept)} \\
\midrule
                 & ald      & 0.865 ± 1.337          & \textbf{0.278 ± 0.008} & 0.653 ± 0.014          & 0.648 ± 0.011          & 0.407 ± 0.343          & 1.025 ± 0.016 & 0.005 ± 0.030 & 1.027 ± 0.042 & -0.016 ± 0.037 \\
                 & CQRNN    & 0.278 ± 0.144          & 0.326 ± 0.034          & \textbf{0.657 ± 0.008} & \textbf{0.651 ± 0.007} & 0.466 ± 0.150          & \textbf{1.001 ± 0.062} & \textbf{-0.003 ± 0.026} & \textbf{1.007 ± 0.039} & -0.020 ± 0.047 \\
Norm\_linear     & LogNorm  & 0.372 ± 0.228          & 0.709 ± 0.028          & 0.652 ± 0.016          & 0.646 ± 0.014          & 0.496 ± 0.399          & 0.965 ± 0.024 & 0.005 ± 0.014 & 0.978 ± 0.041 & 0.014 ± 0.067 \\
                 & DeepSurv & \textbf{0.239 ± 0.114} & 0.676 ± 0.026          & \textbf{0.657 ± 0.008} & \textbf{0.651 ± 0.007} & \textbf{0.139 ± 0.071} & 0.983 ± 0.018 & 0.015 ± 0.016 & \textbf{1.007 ± 0.018} & \textbf{-0.005 ± 0.014} \\
                 & DeepHit  & 1.481 ± 0.527          & 0.503 ± 0.025          & 0.635 ± 0.024          & 0.628 ± 0.025          & 6.540 ± 1.458          & 0.967 ± 0.036 & 0.098 ± 0.070 & 1.216 ± 0.051 & -0.302 ± 0.029 \\
\midrule
                 & ald      & 0.243 ± 0.080          & \textbf{0.212 ± 0.006} & 0.670 ± 0.015          & 0.644 ± 0.016          & 0.406 ± 0.179          & 1.072 ± 0.021 & -0.011 ± 0.015 & 1.038 ± 0.025 & -0.016 ± 0.040 \\
                 & CQRNN    & \textbf{0.117 ± 0.037} & 0.507 ± 0.026          & \textbf{0.674 ± 0.014} & \textbf{0.651 ± 0.014} & 0.241 ± 0.099          & 0.983 ± 0.026 & \textbf{0.002 ± 0.018} & \textbf{0.987 ± 0.012} & 0.011 ± 0.027 \\
Norm\_nonlinear  & LogNorm  & 0.396 ± 0.432          & 0.560 ± 0.058          & 0.630 ± 0.087          & 0.617 ± 0.074          & 2.136 ± 3.886          & 1.003 ± 0.052 & 0.051 ± 0.054 & 1.097 ± 0.060 & -0.098 ± 0.059 \\
                 & DeepSurv & 0.197 ± 0.047          & 0.623 ± 0.013          & 0.670 ± 0.015          & 0.650 ± 0.014          & \textbf{0.196 ± 0.128} & 1.015 ± 0.019 & 0.007 ± 0.016 & 1.019 ± 0.022 & \textbf{-0.007 ± 0.026} \\
                 & DeepHit  & 1.099 ± 0.130          & 0.515 ± 0.049          & 0.610 ± 0.051          & 0.596 ± 0.040          & 3.886 ± 3.682          & \textbf{0.999 ± 0.064} & -0.007 ± 0.061 & 1.064 ± 0.067 & -0.161 ± 0.084 \\
\midrule
                 & ald      & 0.473 ± 0.344          & \textbf{0.045 ± 0.002} & 0.785 ± 0.010          & 0.703 ± 0.019          & \textbf{0.115 ± 0.030} & 1.019 ± 0.020 & \textbf{0.002 ± 0.016} & 1.016 ± 0.015 & \textbf{-0.006 ± 0.021} \\
                 & CQRNN    & \textbf{0.301 ± 0.104} & 0.535 ± 0.015          & \textbf{0.786 ± 0.009} & \textbf{0.706 ± 0.015} & 0.162 ± 0.141          & 1.018 ± 0.033 & -0.013 ± 0.013 & \textbf{1.002 ± 0.015} & -0.007 ± 0.017 \\
Norm uniform     & LogNorm  & 17.079 ± 5.833         & 0.387 ± 0.013          & 0.615 ± 0.118          & 0.578 ± 0.083          & 3.799 ± 0.354          & 0.951 ± 0.059 & 0.159 ± 0.043 & 1.186 ± 0.016 & -0.129 ± 0.021 \\
                 & DeepSurv & 0.627 ± 0.180          & 0.516 ± 0.009          & 0.781 ± 0.014          & 0.701 ± 0.020          & 0.466 ± 0.149          & 1.038 ± 0.017 & 0.022 ± 0.013 & 1.069 ± 0.012 & -0.051 ± 0.016 \\
                 & DeepHit  & 1.468 ± 0.458          & 0.364 ± 0.048          & 0.758 ± 0.033          & 0.688 ± 0.028          & 3.150 ± 1.142          & \textbf{1.015 ± 0.045} & 0.024 ± 0.047 & 1.128 ± 0.051 & -0.209 ± 0.027 \\
\midrule
                 & ald      & 2.942 ± 2.389          & \textbf{0.309 ± 0.018} & \textbf{0.560 ± 0.008} & \textbf{0.560 ± 0.007} & 0.432 ± 0.405          & 0.978 ± 0.047 & -0.015 ± 0.014 & 0.964 ± 0.049 & 0.016 ± 0.053 \\
                 & CQRNN    & 1.943 ± 0.297          & 0.317 ± 0.013          & 0.558 ± 0.013          & 0.557 ± 0.011          & 0.305 ± 0.129          & 0.976 ± 0.066 & 0.012 ± 0.043 & \textbf{1.001 ± 0.027} & \textbf{-0.008 ± 0.019} \\
Exponential      & LogNorm  & 3.223 ± 0.823          & 0.455 ± 0.010          & 0.527 ± 0.028          & 0.528 ± 0.028          & 0.419 ± 0.141          & 0.983 ± 0.026 & 0.042 ± 0.018 & 1.057 ± 0.022 & -0.051 ± 0.021 \\
                 & DeepSurv & \textbf{1.913 ± 0.269} & 0.486 ± 0.015          & 0.558 ± 0.007          & 0.558 ± 0.006          & \textbf{0.119 ± 0.066} & \textbf{0.986 ± 0.033} & \textbf{0.009 ± 0.022} & 1.003 ± 0.018 & \textbf{-0.008 ± 0.018} \\
                 & DeepHit  & 2.626 ± 2.759          & 0.471 ± 0.012          & 0.526 ± 0.032          & 0.526 ± 0.031          & 1.205 ± 1.060          & 0.960 ± 0.027 & -0.012 ± 0.021 & 0.907 ± 0.055 & 0.127 ± 0.066 \\
\midrule
                 & ald      & 5.135 ± 9.533          & \textbf{0.219 ± 0.028} & 0.767 ± 0.009          & 0.763 ± 0.009          & 0.648 ± 0.511          & 1.044 ± 0.023 & -0.023 ± 0.033 & 0.993 ± 0.049 & 0.021 ± 0.060 \\
                 & CQRNN    & \textbf{0.350 ± 0.098} & 0.461 ± 0.030          & \textbf{0.775 ± 0.005} & \textbf{0.769 ± 0.005} & 0.346 ± 0.131          & 0.989 ± 0.057 & \textbf{-0.001 ± 0.042} &\textbf{0.995 ± 0.022} & \textbf{-0.003 ± 0.023} \\
Weibull          & LogNorm  & 0.862 ± 0.121          & 0.840 ± 0.021          & 0.773 ± 0.006          & 0.767 ± 0.006          & 0.598 ± 0.172          & 0.993 ± 0.026 & 0.029 ± 0.016 & 1.050 ± 0.028 & -0.053 ± 0.038 \\
                 & DeepSurv & 0.381 ± 0.098          & 0.969 ± 0.019          & 0.772 ± 0.004          & 0.766 ± 0.006          & \textbf{0.118 ± 0.049} & 0.989 ± 0.023 & 0.006 ± 0.012 & \textbf{1.005 ± 0.018} & -0.009 ± 0.021 \\
                 & DeepHit  & 1.975 ± 0.172          & 0.618 ± 0.032          & 0.769 ± 0.006          & 0.763 ± 0.007          & 3.020 ± 1.157          & \textbf{0.998 ± 0.058} & 0.122 ± 0.071 & 1.206 ± 0.056 & -0.187 ± 0.069 \\
\midrule
                 & ald      & 0.363 ± 0.068          & \textbf{0.376 ± 0.013} & 0.588 ± 0.014          & 0.585 ± 0.014          & 0.256 ± 0.150          & 1.005 ± 0.021 & 0.006 ± 0.011 & 1.011 ± 0.028 & -0.004 ± 0.029 \\
                 & CQRNN    & 0.950 ± 0.091          & 0.407 ± 0.019          & 0.588 ± 0.016          & 0.584 ± 0.016          & 0.459 ± 0.220          & 1.024 ± 0.066 & -0.019 ± 0.034 & 0.996 ± 0.042 & \textbf{0.000 ± 0.031} \\
LogNorm          & LogNorm  & \textbf{0.267 ± 0.062} & 0.645 ± 0.021          & 0.588 ± 0.015          & 0.584 ± 0.015          & \textbf{0.103 ± 0.020} & 1.009 ± 0.012 & 0.006 ± 0.009 & 1.016 ± 0.015 & -0.010 ± 0.017 \\
                 & DeepSurv & 0.963 ± 0.058          & 0.658 ± 0.029          & \textbf{0.589 ± 0.016} & \textbf{0.586 ± 0.016} & 0.137 ± 0.049 & \textbf{0.996 ± 0.021} & \textbf{0.001 ± 0.020} & \textbf{0.997 ± 0.025} & 0.002 ± 0.021 \\
                 & DeepHit  & 0.902 ± 0.504          & 0.568 ± 0.025          & 0.551 ± 0.032          & 0.548 ± 0.031          & 2.088 ± 1.666          & 0.988 ± 0.031 & -0.026 ± 0.050 & 0.892 ± 0.072 & 0.162 ± 0.090 \\
\midrule
                 & ald      & 0.667 ± 0.139          & \textbf{0.019 ± 0.001} & \textbf{0.919 ± 0.007} & \textbf{0.870 ± 0.029} & \textbf{0.036 ± 0.017} & 1.009 ± 0.005 & -0.004 ± 0.004 & 1.001 ± 0.009 & \textbf{-0.002 ± 0.010} \\
                 & CQRNN    & \textbf{0.574 ± 0.031} & 0.538 ± 0.006          & 0.914 ± 0.008          & 0.863 ± 0.033          & 0.062 ± 0.099          & \textbf{1.000 ± 0.019} & -0.002 ± 0.007 & \textbf{1.000 ± 0.012} & -0.004 ± 0.011 \\
Norm heavy       & LogNorm  & 33.140 ± 12.004        & 0.411 ± 0.014          & 0.781 ± 0.071          & 0.679 ± 0.126          & 2.249 ± 0.490          & 1.122 ± 0.022 & \textbf{0.001 ± 0.029} & 1.111 ± 0.031 & -0.074 ± 0.032 \\
                 & DeepSurv & 1.662 ± 0.157          & 0.558 ± 0.007          & 0.726 ± 0.035          & 0.582 ± 0.056          & 0.577 ± 0.067          & 1.070 ± 0.006 & -0.002 ± 0.009 & 1.065 ± 0.011 & -0.065 ± 0.010 \\
                 & DeepHit  & 0.814 ± 0.104          & 0.475 ± 0.037          & 0.913 ± 0.009          & 0.856 ± 0.034          & 1.349 ± 0.374          & 1.051 ± 0.044 & 0.055 ± 0.035 & 1.139 ± 0.027 & -0.121 ± 0.036 \\
\midrule
                 & ald      & \textbf{0.238 ± 0.036} & \textbf{0.047 ± 0.003} & \textbf{0.894 ± 0.005} & \textbf{0.872 ± 0.004} & 0.157 ± 0.044          & 1.058 ± 0.012 & -0.035 ± 0.011 & \textbf{0.997 ± 0.012} & \textbf{0.004 ± 0.014} \\
                 & CQRNN    & 0.312 ± 0.033          & 0.608 ± 0.010          & 0.888 ± 0.005          & 0.867 ± 0.005          & \textbf{0.097 ± 0.045} & \textbf{0.984 ± 0.026} & \textbf{0.001 ± 0.013} & 0.989 ± 0.019 & 0.007 ± 0.020 \\
Norm med.        & LogNorm  & 7.300 ± 2.579          & 0.430 ± 0.019          & 0.810 ± 0.048          & 0.777 ± 0.048          & 8.192 ± 0.660          & 0.751 ± 0.073 & 0.350 ± 0.052 & 1.280 ± 0.021 & -0.192 ± 0.036 \\
                 & DeepSurv & 0.253 ± 0.026          & 0.722 ± 0.012          & 0.893 ± 0.004          & 0.871 ± 0.004          & 0.609 ± 0.111          & 1.054 ± 0.014 & 0.008 ± 0.015 & 1.061 ± 0.019 & -0.051 ± 0.017 \\
                 & DeepHit  & 0.916 ± 0.077          & 0.576 ± 0.012          & 0.886 ± 0.006          & 0.863 ± 0.005          & 1.655 ± 0.409          & 1.056 ± 0.032 & 0.038 ± 0.026 & 1.130 ± 0.032 & -0.130 ± 0.066 \\
\midrule
                 & ald      & \textbf{0.236 ± 0.051} & \textbf{0.090 ± 0.007} & \textbf{0.882 ± 0.004} & \textbf{0.874 ± 0.004} & 0.339 ± 0.076          & 1.087 ± 0.017 & -0.050 ± 0.011 & \textbf{0.999 ± 0.014} & 0.005 ± 0.021 \\
                 & CQRNN    & 0.271 ± 0.032          & 0.671 ± 0.013          & 0.879 ± 0.002          & 0.871 ± 0.002          & \textbf{0.149 ± 0.097} & \textbf{0.999 ± 0.027} & \textbf{-0.014 ± 0.017} & 0.985 ± 0.022 & \textbf{0.004 ± 0.027} \\
Norm light       & LogNorm  & 3.152 ± 2.154          & 0.548 ± 0.023          & 0.832 ± 0.022          & 0.821 ± 0.022          & 12.884 ± 1.700         & 0.804 ± 0.162 & 0.358 ± 0.126 & 1.351 ± 0.018 & -0.256 ± 0.038 \\
                 & DeepSurv & 0.247 ± 0.016          & 0.941 ± 0.024          & \textbf{0.882 ± 0.002} & \textbf{0.874 ± 0.002} & 0.582 ± 0.127          & 1.038 ± 0.014 & 0.016 ± 0.019 & 1.057 ± 0.023 & -0.040 ± 0.023 \\
                 & DeepHit  & 0.959 ± 0.051          & 0.691 ± 0.030          & 0.875 ± 0.004          & 0.867 ± 0.004          & 1.854 ± 0.461          & 1.044 ± 0.041 & 0.063 ± 0.022 & 1.159 ± 0.023 & -0.157 ± 0.053 \\
\midrule
                 & ald      & 0.405 ± 0.079          & \textbf{0.066 ± 0.003} & 0.890 ± 0.005          & 0.847 ± 0.008          & \textbf{0.114 ± 0.036} & 1.007 ± 0.014 & 0.006 ± 0.012 & 1.004 ± 0.018 & 0.010 ± 0.022 \\
                 & CQRNN    & 0.301 ± 0.024          & 0.568 ± 0.018          & 0.886 ± 0.006          & 0.841 ± 0.016          & 0.147 ± 0.109          & 0.988 ± 0.028 & \textbf{0.003 ± 0.014} & \textbf{0.999 ± 0.024} & \textbf{-0.007 ± 0.031} \\
Norm same        & LogNorm  & 0.379 ± 0.202          & 0.770 ± 0.039          & \textbf{0.894 ± 0.005} & \textbf{0.850 ± 0.009} & 0.900 ± 0.801          & \textbf{0.994 ± 0.032} & 0.017 ± 0.030 & 1.036 ± 0.087 & -0.048 ± 0.110 \\
                 & DeepSurv & \textbf{0.254 ± 0.036} & 0.787 ± 0.015          & 0.889 ± 0.004          & 0.837 ± 0.025          & 0.227 ± 0.053          & 1.033 ± 0.014 & -0.014 ± 0.011 & 1.010 ± 0.016 & -0.010 ± 0.018 \\
                 & DeepHit  & 1.303 ± 0.132          & 0.572 ± 0.032          & 0.882 ± 0.006          & 0.832 ± 0.017          & 1.798 ± 0.770          & 1.041 ± 0.049 & 0.060 ± 0.047 & 1.142 ± 0.041 & -0.124 ± 0.059 \\
\midrule
                 & ald      & \textbf{0.385 ± 0.193} & \textbf{0.095 ± 0.006} & \textbf{0.777 ± 0.012} & \textbf{0.727 ± 0.021} & \textbf{0.043 ± 0.019} & \textbf{1.003 ± 0.014} & -0.005 ± 0.005 & \textbf{0.998 ± 0.014} & \textbf{-0.003 ± 0.014} \\
                 & CQRNN    & 0.717 ± 0.027          & 0.436 ± 0.035          & 0.767 ± 0.009          & 0.718 ± 0.018          & 0.235 ± 0.104          & 0.992 ± 0.026 & -0.007 ± 0.013 & \textbf{0.998 ± 0.032} & -0.019 ± 0.035 \\
LogNorm heavy    & LogNorm  & 0.755 ± 0.194          & 0.401 ± 0.012          & 0.643 ± 0.053          & 0.609 ± 0.046          & 0.066 ± 0.056          & 1.018 ± 0.012 & \textbf{-0.002 ± 0.005} & 1.011 ± 0.019 & \textbf{-0.003 ± 0.020} \\
                 & DeepSurv & 0.842 ± 0.019          & 0.459 ± 0.013          & 0.497 ± 0.034          & 0.465 ± 0.029          & 0.102 ± 0.068          & 1.031 ± 0.008 & -0.010 ± 0.007 & 1.015 ± 0.017 & -0.018 ± 0.015 \\
                 & DeepHit  & 0.724 ± 0.020          & 0.402 ± 0.012          & 0.756 ± 0.012          & 0.712 ± 0.019          & 0.282 ± 0.121          & 1.036 ± 0.020 & -0.012 ± 0.006 & 1.030 ± 0.014 & -0.045 ± 0.018 \\
\midrule
                 & ald      & \textbf{0.178 ± 0.046} & \textbf{0.174 ± 0.005} & \textbf{0.747 ± 0.004} & \textbf{0.718 ± 0.007} & 0.087 ± 0.052          & 1.008 ± 0.017 & -0.004 ± 0.010 & \textbf{1.002 ± 0.009} & -0.001 ± 0.012 \\
                 & CQRNN    & 0.540 ± 0.059          & 0.368 ± 0.053          & 0.746 ± 0.005          & 0.716 ± 0.006          & 0.376 ± 0.166          & 0.985 ± 0.069 & \textbf{-0.001 ± 0.037} & 0.994 ± 0.035 & -0.006 ± 0.041 \\
LogNorm med.     & LogNorm  & 0.549 ± 0.101          & 0.452 ± 0.012          & 0.694 ± 0.024          & 0.665 ± 0.021          & \textbf{0.085 ± 0.067} & \textbf{1.002 ± 0.016} & 0.007 ± 0.011 & 1.008 ± 0.022 & \textbf{0.000 ± 0.026} \\
                 & DeepSurv & 0.654 ± 0.029          & 0.545 ± 0.015          & 0.638 ± 0.011          & 0.596 ± 0.011          & 0.138 ± 0.058          & 1.020 ± 0.017 & -0.015 ± 0.009 & 0.994 ± 0.016 & 0.005 ± 0.018 \\
                 & DeepHit  & 0.600 ± 0.018          & 0.426 ± 0.010          & 0.729 ± 0.019          & 0.702 ± 0.015          & 0.344 ± 0.118          & 1.046 ± 0.018 & -0.032 ± 0.017 & 0.986 ± 0.018 & 0.018 ± 0.020 \\
\midrule
                 & ald      & \textbf{0.184 ± 0.035} & \textbf{0.310 ± 0.011} & \textbf{0.725 ± 0.007} & 0.713 ± 0.008          & \textbf{0.185 ± 0.095} & 0.985 ± 0.015 & 0.007 ± 0.009 & \textbf{1.001 ± 0.014} & \textbf{-0.001 ± 0.017} \\
                 & CQRNN    & 0.356 ± 0.073          & 0.418 ± 0.045          & \textbf{0.725 ± 0.007} & \textbf{0.714 ± 0.008} & 0.976 ± 0.602          & 0.988 ± 0.077 & -0.012 ± 0.045 & 0.962 ± 0.071 & 0.044 ± 0.072 \\
LogNorm light    & LogNorm  & 0.311 ± 0.022          & 0.794 ± 0.026          & 0.709 ± 0.009          & 0.698 ± 0.010          & 0.231 ± 0.170          & 0.972 ± 0.027 & -0.007 ± 0.014 & 0.964 ± 0.035 & 0.029 ± 0.041 \\
                 & DeepSurv & 0.403 ± 0.027          & 0.833 ± 0.025          & 0.715 ± 0.009          & 0.700 ± 0.011          & 0.211 ± 0.123          & 1.010 ± 0.017 & \textbf{-0.000 ± 0.012} & 1.004 ± 0.017 & 0.005 ± 0.018 \\
                 & DeepHit  & 0.581 ± 0.018          & 0.654 ± 0.017          & 0.702 ± 0.008          & 0.692 ± 0.008          & 0.253 ± 0.174          & \textbf{1.006 ± 0.030} & -0.013 ± 0.016 & 0.974 ± 0.021 & 0.042 ± 0.026 \\
\midrule
                 & ald      & \textbf{0.191 ± 0.044} & \textbf{0.154 ± 0.006} & 0.739 ± 0.009          & 0.697 ± 0.008          & \textbf{0.076 ± 0.057} & \textbf{1.012 ± 0.011} & \textbf{-0.001 ± 0.008} & 1.009 ± 0.011 & -0.005 ± 0.010 \\
                 & CQRNN    & 0.319 ± 0.079          & 0.300 ± 0.049          & 0.740 ± 0.008          & 0.698 ± 0.009          & 0.787 ± 0.336          & 0.986 ± 0.086 & -0.003 ± 0.040 & 0.971 ± 0.058 & 0.041 ± 0.056 \\
LogNorm same     & LogNorm  & 0.273 ± 0.068          & 0.528 ± 0.017          & 0.736 ± 0.012          & 0.695 ± 0.010          & 0.213 ± 0.117          & 0.972 ± 0.015 & -0.006 ± 0.010 & 0.963 ± 0.028 & 0.033 ± 0.040 \\
                 & DeepSurv & 0.362 ± 0.026          & 0.511 ± 0.012          & \textbf{0.743 ± 0.010} & \textbf{0.700 ± 0.007} & 0.138 ± 0.040          & 1.017 ± 0.013 & -0.005 ± 0.012 & \textbf{1.004 ± 0.014} & \textbf{0.001 ± 0.013} \\
                 & DeepHit  & 0.560 ± 0.098          & 0.385 ± 0.022          & 0.652 ± 0.066          & 0.633 ± 0.047          & 1.265 ± 1.911          & 0.925 ± 0.071 & -0.010 ± 0.011 & 0.925 ± 0.088 & 0.058 ± 0.108 \\
\midrule
                 & ald      & 1.626 ± 0.194          & \textbf{0.245 ± 0.012} & 0.637 ± 0.021          & 0.633 ± 0.031          & 0.293 ± 0.125          & \textbf{1.001 ± 0.033} & -0.011 ± 0.016 & 0.993 ± 0.028 & \textbf{-0.009 ± 0.024} \\
                 & CQRNN    & 0.998 ± 0.074          & 0.344 ± 0.027          & 0.632 ± 0.017          & 0.630 ± 0.033          & 0.641 ± 0.391          & 0.972 ± 0.048 & \textbf{0.007 ± 0.018} & \textbf{1.001 ± 0.042} & -0.022 ± 0.051 \\
METABRIC         & LogNorm  & 1.329 ± 0.041          & 0.526 ± 0.015          & 0.609 ± 0.019          & 0.613 ± 0.046          & 0.619 ± 0.247          & 0.964 ± 0.026 & -0.024 ± 0.011 & 0.937 ± 0.017 & 0.040 ± 0.018 \\
                 & DeepSurv & \textbf{0.981 ± 0.029} & 0.533 ± 0.019          & \textbf{0.645 ± 0.016} & \textbf{0.635 ± 0.035} & \textbf{0.159 ± 0.075} & 1.009 ± 0.011 & -0.008 ± 0.013 & 1.003 ± 0.022 & -0.010 ± 0.023 \\
                 & DeepHit  & 1.177 ± 0.065          & 0.462 ± 0.008          & 0.563 ± 0.040          & 0.577 ± 0.053          & 0.659 ± 0.212          & 1.070 ± 0.022 & -0.036 ± 0.014 & 1.018 ± 0.015 & -0.024 ± 0.029 \\
\bottomrule
\end{tabular}
}
\end{table}

\begin{table}[t]
\centering
\resizebox{1\textwidth}{!}{
\begin{tabular}{ccccccccccc}
\toprule
\textbf{Dataset} & \textbf{Method} & \textbf{MAE} & \textbf{IBS} & \textbf{Harrell’s C-index} & \textbf{Uno’s C-index} & \textbf{CensDcal} & \textbf{Cal$[S(t|\mathbf{x})]$(Slope)} & \textbf{Cal$[S(t|\mathbf{x})]$(Intercept)} & \textbf{Cal$[f(t|\mathbf{x})]$(Slope)} & \textbf{Cal$[f(t|\mathbf{x})]$(Intercept)} \\
\midrule
                 & ald      & 2.196 ± 0.612          & \textbf{0.134 ± 0.013} & 0.823 ± 0.016          & 0.824 ± 0.014          & \textbf{0.198 ± 0.094} & 0.972 ± 0.027          & 0.003 ± 0.016          & 0.981 ± 0.021 & 0.009 ± 0.023 \\
                 & CQRNN    & \textbf{0.798 ± 0.049} & 0.636 ± 0.018          & \textbf{0.838 ± 0.016} & \textbf{0.846 ± 0.016} & 0.564 ± 0.248          & 0.974 ± 0.060          & \textbf{0.002 ± 0.022} & \textbf{0.998 ± 0.057} & -0.024 ± 0.053 \\
WHAS             & LogNorm  & 1.976 ± 0.232          & 0.614 ± 0.025          & 0.600 ± 0.042          & 0.575 ± 0.039          & 0.584 ± 0.233          & 0.920 ± 0.032          & 0.041 ± 0.021          & 0.994 ± 0.032 & \textbf{-0.002 ± 0.033} \\
                 & DeepSurv & 0.867 ± 0.050          & 0.699 ± 0.023          & 0.711 ± 0.014          & 0.637 ± 0.025          & 0.228 ± 0.101          & \textbf{0.997 ± 0.020} & 0.005 ± 0.018          & 1.012 ± 0.020 & -0.019 ± 0.019 \\
                 & DeepHit  & 0.966 ± 0.077          & 0.604 ± 0.023          & 0.806 ± 0.018          & 0.811 ± 0.018          & 0.269 ± 0.172          & 0.963 ± 0.036          & 0.018 ± 0.022          & 1.008 ± 0.026 & -0.015 ± 0.031 \\
\midrule
                 & ald      & 1.121 ± 0.107          & 0.362 ± 0.013          & 0.568 ± 0.015          & 0.572 ± 0.015          & 2.197 ± 0.667          & 1.084 ± 0.043          & -0.113 ± 0.023         & 0.900 ± 0.056 & 0.084 ± 0.046 \\
                 & CQRNN    & 0.659 ± 0.047          & \textbf{0.344 ± 0.007} & \textbf{0.612 ± 0.005} & \textbf{0.613 ± 0.006} & 0.724 ± 0.428          & 1.034 ± 0.066          & -0.019 ± 0.034         & \textbf{0.992 ± 0.051} & 0.019 ± 0.049 \\
SUPPORT          & LogNorm  & 1.311 ± 0.150          & 0.688 ± 0.020          & 0.597 ± 0.011          & 0.597 ± 0.011          & 2.792 ± 0.942          & 0.942 ± 0.040          & -0.114 ± 0.008         & 0.769 ± 0.041 & 0.169 ± 0.042 \\
                 & DeepSurv & \textbf{0.511 ± 0.021} & 0.629 ± 0.014          & 0.599 ± 0.008          & 0.597 ± 0.009          & \textbf{0.092 ± 0.036} & \textbf{0.989 ± 0.016} & \textbf{-0.003 ± 0.011}& 0.986 ± 0.012 & \textbf{0.006 ± 0.014} \\
                 & DeepHit  & 0.574 ± 0.034          & 0.530 ± 0.009          & 0.577 ± 0.008          & 0.582 ± 0.009          & 0.829 ± 0.213          & 0.891 ± 0.013          & 0.006 ± 0.008          & 0.909 ± 0.014  & 0.086 ± 0.021 \\
\midrule
                 & ald      & 1.713 ± 0.208          & \textbf{0.279 ± 0.014} & 0.671 ± 0.013          & 0.665 ± 0.013          & 0.283 ± 0.106          & \textbf{1.000 ± 0.035} & -0.018 ± 0.016         & 0.977 ± 0.025 & 0.014 ± 0.034 \\
                 & CQRNN    & 0.865 ± 0.070          & 0.357 ± 0.021          & \textbf{0.680 ± 0.015} & \textbf{0.672 ± 0.014} & 0.573 ± 0.577          & 0.953 ± 0.043          & -0.008 ± 0.016         & 0.967 ± 0.030 & \textbf{0.002 ± 0.040} \\
GBSG             & LogNorm  & 1.469 ± 0.105          & 0.577 ± 0.015          & 0.660 ± 0.012          & 0.653 ± 0.012          & 0.817 ± 0.303          & 0.968 ± 0.025          & -0.057 ± 0.011         & 0.886 ± 0.025 & 0.086 ± 0.035 \\
                 & DeepSurv & \textbf{0.709 ± 0.036} & 0.569 ± 0.016          & 0.611 ± 0.017          & 0.602 ± 0.016          & \textbf{0.180 ± 0.126} & 1.002 ± 0.021          & \textbf{-0.003 ± 0.013}& \textbf{0.996 ± 0.013} & 0.004 ± 0.018 \\
                 & DeepHit  & 0.773 ± 0.037          & 0.495 ± 0.016          & 0.649 ± 0.016          & 0.644 ± 0.016          & 2.020 ± 1.450          & 0.967 ± 0.049          & -0.045 ± 0.014         & 0.952 ± 0.031 & -0.025 ± 0.016 \\
\midrule
                 & ald      & 3.002 ± 1.497          & \textbf{0.245 ± 0.015} & 0.561 ± 0.037          & 0.547 ± 0.040          & 0.835 ± 0.604          & 1.053 ± 0.045          & -0.038 ± 0.025         & \textbf{0.994 ± 0.021} & \textbf{0.004 ± 0.025} \\
                 & CQRNN    & 1.008 ± 0.053          & 0.272 ± 0.013          & \textbf{0.567 ± 0.022} & \textbf{0.557 ± 0.017} & 0.251 ± 0.123          & 0.967 ± 0.037          & 0.011 ± 0.026          & 0.988 ± 0.027 & 0.009 ± 0.020 \\
TMBImmuno        & LogNorm  & 1.880 ± 0.156          & 0.420 ± 0.011          & 0.561 ± 0.028          & \textbf{0.557 ± 0.028} & 0.617 ± 0.196          & 0.949 ± 0.028          & -0.027 ± 0.019         & 0.913 ± 0.025 & 0.066 ± 0.027 \\
                 & DeepSurv & \textbf{0.948 ± 0.097} & 0.395 ± 0.012          & 0.543 ± 0.034          & 0.526 ± 0.039          & \textbf{0.246 ± 0.168} & \textbf{1.019 ± 0.030} & \textbf{-0.001 ± 0.023}& 1.009 ± 0.020 & -0.003 ± 0.018 \\
                 & DeepHit  & 1.117 ± 0.141          & 0.400 ± 0.011          & 0.560 ± 0.023          & 0.554 ± 0.021          & 0.464 ± 0.214          & 0.963 ± 0.039          & -0.018 ± 0.026         & 0.935 ± 0.020 & 0.058 ± 0.026 \\
\midrule
                 & ald      & 2.593 ± 0.289          & \textbf{0.086 ± 0.008} & \textbf{0.617 ± 0.032} & 0.568 ± 0.036          & \textbf{0.066 ± 0.027} & \textbf{1.002 ± 0.019} & -0.007 ± 0.010         & \textbf{0.993 ± 0.020} & \textbf{0.003 ± 0.021} \\
                 & CQRNN    & 1.864 ± 0.354          & 0.316 ± 0.035          & 0.599 ± 0.044          & 0.561 ± 0.036          & 0.172 ± 0.083          & 0.993 ± 0.036          & -0.005 ± 0.013         & 0.990 ± 0.030 & \textbf{0.003 ± 0.034} \\
BreastMSK        & LogNorm  & 6.675 ± 0.597          & 0.310 ± 0.015          & 0.610 ± 0.029          & 0.573 ± 0.046          & 0.208 ± 0.089          & 1.044 ± 0.010          & \textbf{-0.004 ± 0.009}& 1.031 ± 0.012 & -0.023 ± 0.015 \\
                 & DeepSurv & 1.639 ± 0.217          & 0.334 ± 0.018          & 0.614 ± 0.033          & \textbf{0.582 ± 0.049} & 0.212 ± 0.099          & 1.046 ± 0.024          & -0.006 ± 0.011         & 1.036 ± 0.019 & -0.036 ± 0.019 \\
                 & DeepHit  & \textbf{1.523 ± 0.076} & 0.303 ± 0.016          & 0.614 ± 0.036          & 0.563 ± 0.046          & 0.411 ± 0.213          & 1.062 ± 0.011          & -0.021 ± 0.006         & 1.032 ± 0.011 & -0.040 ± 0.014 \\
\midrule
                 & ald      & 1.232 ± 0.325          & \textbf{0.108 ± 0.011} & 0.778 ± 0.021          & 0.736 ± 0.030          & 0.450 ± 0.267          & \textbf{0.995 ± 0.047} & 0.003 ± 0.022          & \textbf{0.996 ± 0.038} & \textbf{0.009 ± 0.040} \\
                 & CQRNN    & 0.808 ± 0.197          & 0.375 ± 0.041          & 0.790 ± 0.024          & 0.754 ± 0.034          & 0.543 ± 0.273          & 0.989 ± 0.071          & \textbf{0.001 ± 0.037} & 0.990 ± 0.052 & 0.011 ± 0.058 \\
LGGGBM           & LogNorm  & 1.191 ± 0.214          & 0.382 ± 0.017          & \textbf{0.795 ± 0.022} & \textbf{0.758 ± 0.037} & \textbf{0.327 ± 0.190} & \textbf{1.005 ± 0.025} & 0.007 ± 0.026          & 1.020 ± 0.040 & -0.018 ± 0.044 \\
                 & DeepSurv & \textbf{0.785 ± 0.155} & 0.472 ± 0.024          & 0.728 ± 0.057          & 0.664 ± 0.079          & 0.481 ± 0.219          & 1.022 ± 0.027          & 0.002 ± 0.025          & 1.018 ± 0.040 & -0.012 ± 0.046 \\
                 & DeepHit  & 2.062 ± 0.285          & 0.377 ± 0.024          & 0.769 ± 0.022          & 0.734 ± 0.035          & 1.176 ± 0.539          & 1.085 ± 0.034          & -0.052 ± 0.023         & 0.968 ± 0.035 & 0.066 ± 0.037 \\
\bottomrule
\end{tabular}
}
\end{table}

\begin{figure*}[ht] 
    \centering
    \includegraphics[width=\linewidth]{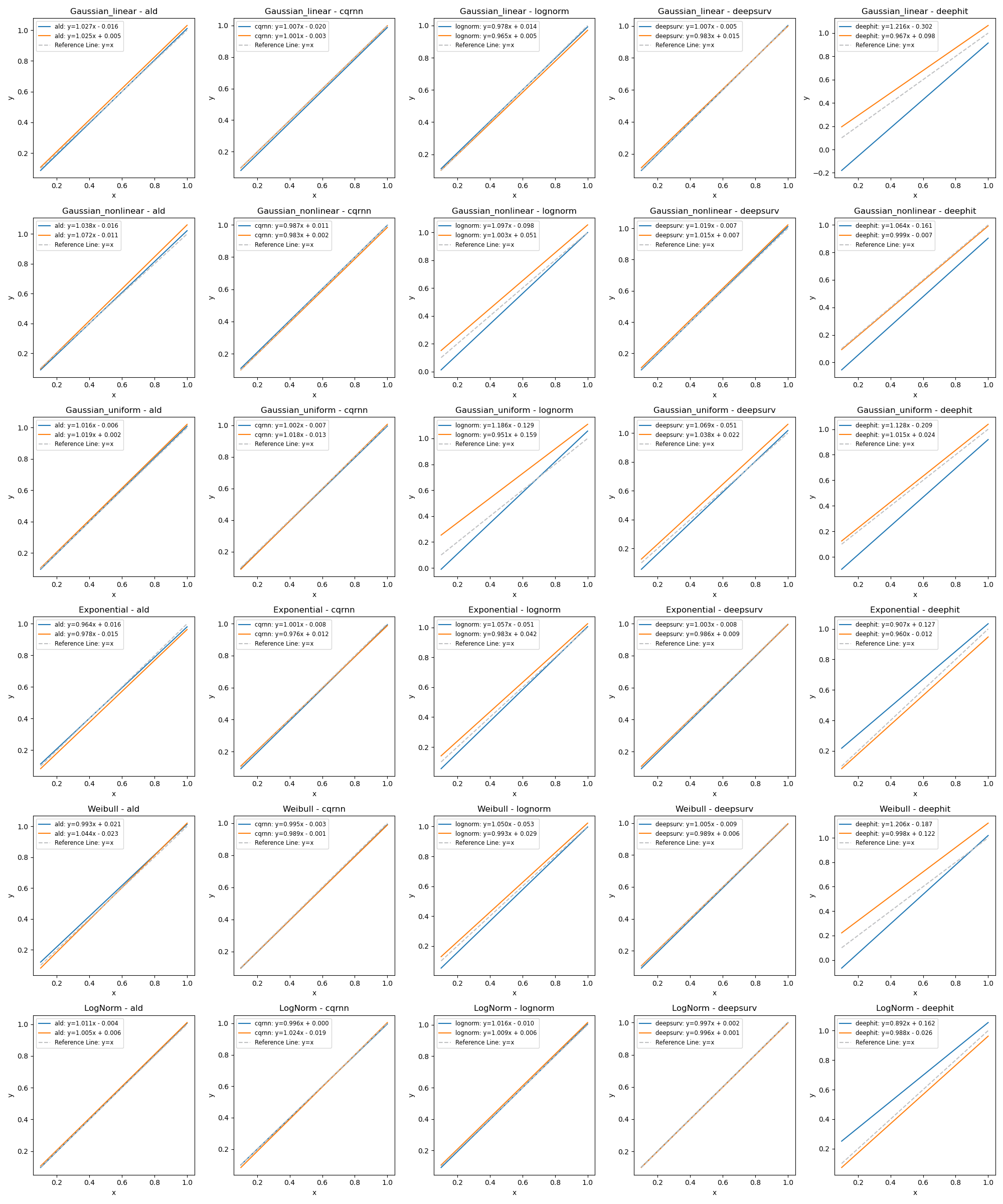} 
\end{figure*}

\begin{figure*}[ht] 
    \centering
    \includegraphics[width=\linewidth]{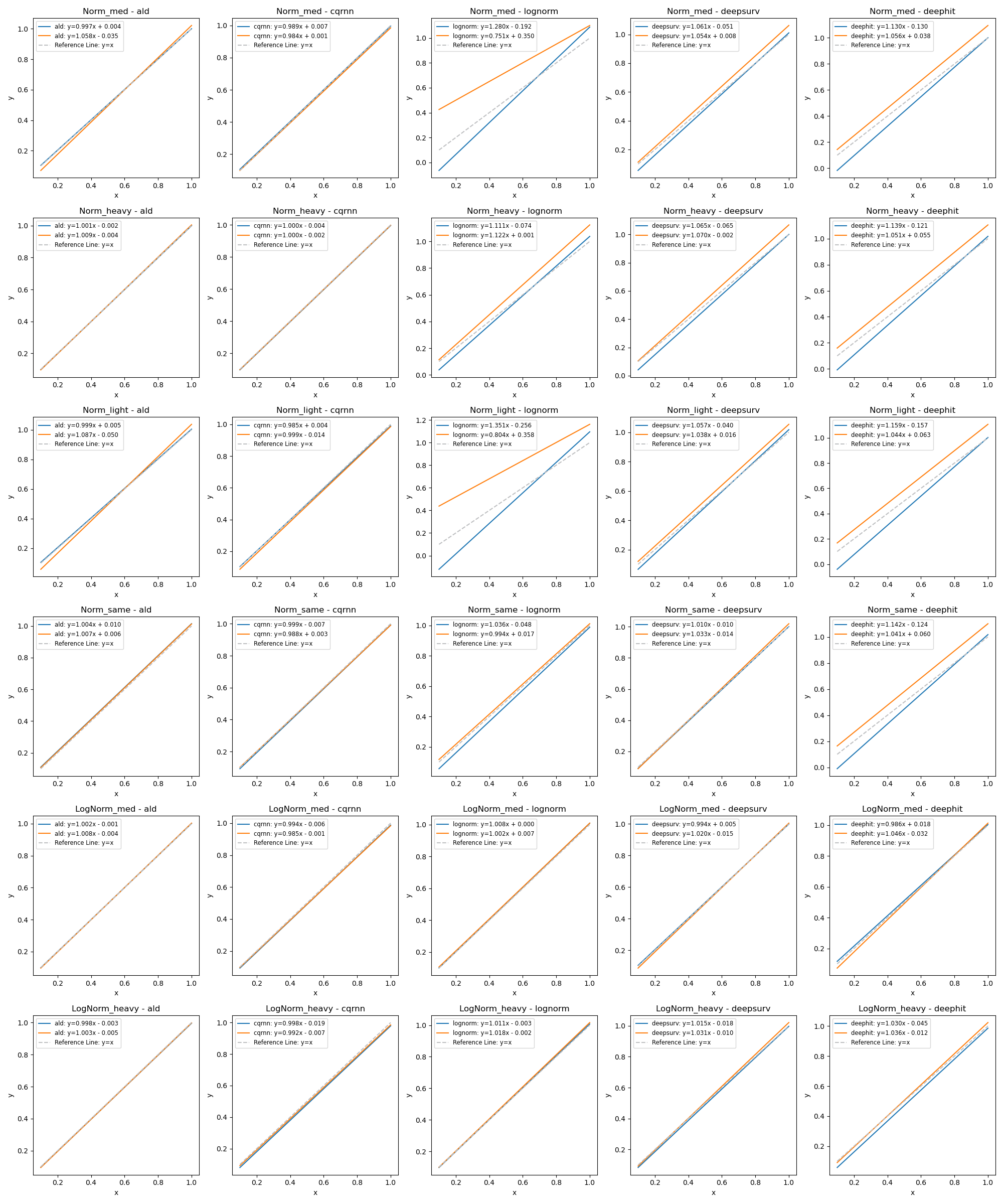} 
    \label{fig:lineplot_all2} 
\end{figure*}

\begin{figure*}[ht] 
    \centering
    \includegraphics[width=\linewidth]{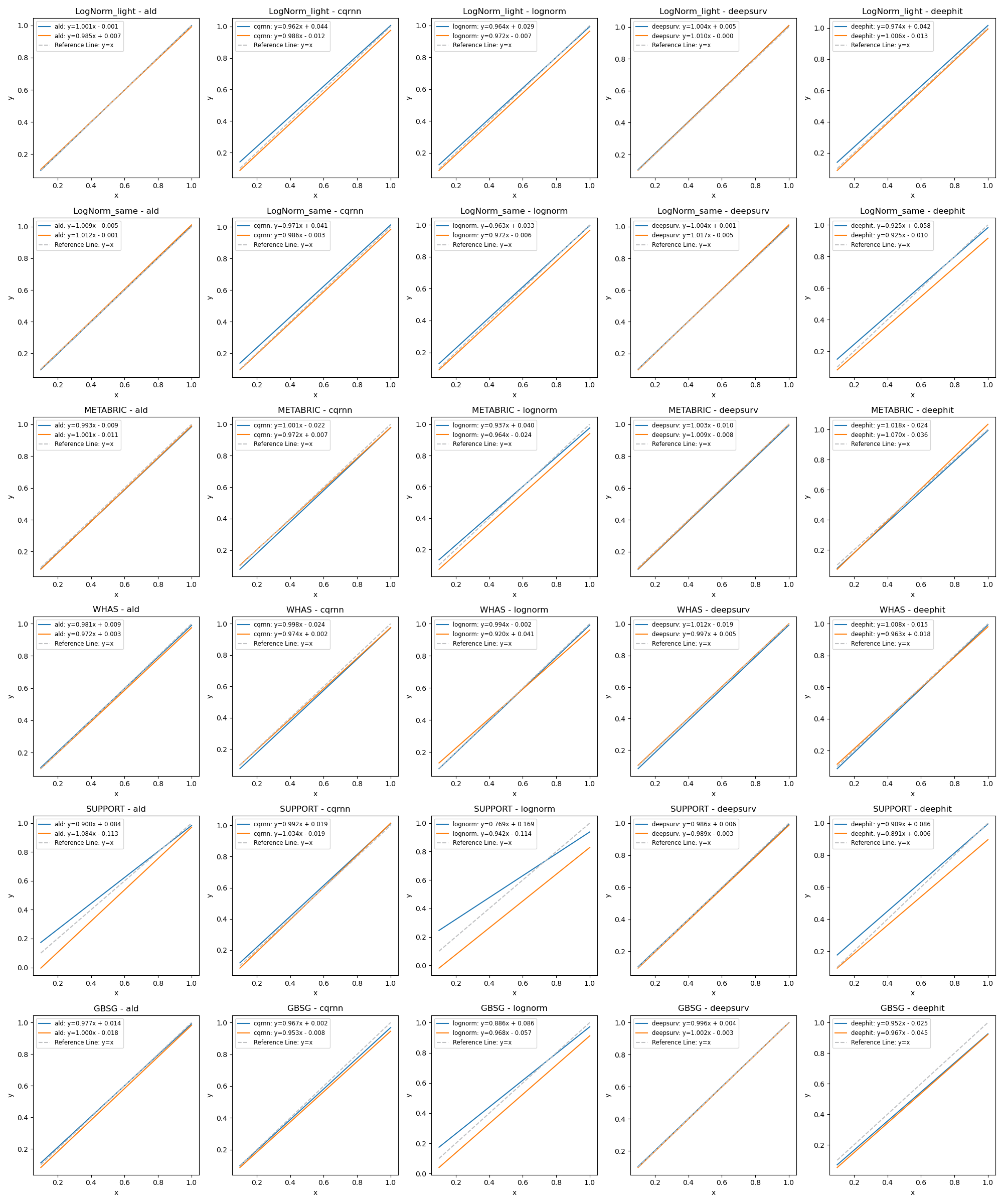} 
    \label{fig:lineplot_all3} 
\end{figure*}

\begin{table}[ht]
\caption{Full results table for all datasets, the ALD method (Mean, Median, Mode), and metrics. The values represent the mean ± 1 standard error on the test set over 10 runs.}
\label{tab:overall results3}
\centering
\resizebox{1\textwidth}{!}{
\begin{tabular}{ccccccccccc}
\toprule
\textbf{Dataset} & \textbf{Method} & \textbf{MAE} & \textbf{IBS} & \textbf{Harrell’s C-index} & \textbf{Uno’s C-index} & \textbf{CensDcal} & \textbf{Cal$[S(t|\mathbf{x})]$(Slope)} & \textbf{Cal$[S(t|\mathbf{x})]$(Intercept)} & \textbf{Cal$[f(t|\mathbf{x})]$(Slope)} & \textbf{Cal$[f(t|\mathbf{x})]$(Intercept)} \\
\midrule
              & ald (Mean)    & 0.865 ± 1.336 &               & 0.653 ± 0.014 & 0.648 ± 0.011 &               &               &                &               &               \\
Norm linear   & ald (Median)  & \textbf{0.217 ± 0.037} & 0.278 ± 0.008 & 0.654 ± 0.012 & 0.682 ± 0.037 & 0.407 ± 0.343 & 1.027 ± 0.042 & -0.016 ± 0.037 & 1.025 ± 0.016 & 0.005 ± 0.030 \\
              & ald (Mode)    & 0.689 ± 0.186 &               & \textbf{0.657 ± 0.008} & \textbf{0.718 ± 0.006}  &               &               &                &               &               \\
\midrule
              & ald (Mean)    & \textbf{0.243 ± 0.080}  &               & \textbf{0.670 ± 0.015} & \textbf{0.644 ± 0.016} &               &               &                &               &               \\
Norm non-lin  & ald (Median)  & 0.253 ± 0.073 & 0.212 ± 0.006 & 0.667 ± 0.015 & 0.582 ± 0.029 & 0.406 ± 0.179 & 1.038 ± 0.025 & -0.016 ± 0.040 & 1.072 ± 0.021 & -0.011 ± 0.015 \\
              & ald (Mode)    & 0.438 ± 0.089 &               & 0.632 ± 0.060 & 0.573 ± 0.054 &               &               &                &               &               \\
\midrule
              & ald (Mean)    & 0.473 ± 0.344 &               & 0.785 ± 0.010 &\textbf{ 0.703 ± 0.020} &               &               &                &               &               \\
Norm uniform  & ald (Median)  & \textbf{0.392 ± 0.196} & 0.045 ± 0.002 & 0.785 ± 0.011 & 0.696 ± 0.013 & 0.115 ± 0.030 & 1.016 ± 0.014 & -0.006 ± 0.021 & 1.019 ± 0.020 & 0.002 ± 0.016 \\
              & ald (Mode)    & 0.613 ± 0.118 &               & \textbf{0.788 ± 0.012} & 0.696 ± 0.014 &               &               &                &               &               \\
\midrule
              & ald (Mean)    & 2.942 ± 2.389 &               & \textbf{0.560 ± 0.008} & \textbf{0.560 ± 0.007} &               &               &                &               &               \\
Exponential   & ald (Median)  & \textbf{1.088 ± 0.308} & 0.309 ± 0.018 & 0.559 ± 0.010 & 0.553 ± 0.020 & 0.432 ± 0.405 & 0.964 ± 0.049 & 0.016 ± 0.053 & 0.978 ± 0.047 & -0.015 ± 0.014 \\
              & ald (Mode)    & 5.009 ± 0.235 &               & 0.556 ± 0.011 & 0.555 ± 0.020 &               &               &                &               &               \\
\midrule
              & ald (Mean)    & 5.134 ± 9.533 &               & \textbf{0.768 ± 0.009} & \textbf{0.763 ± 0.010} &               &               &                &               &               \\
Weibull       & ald (Median)  & \textbf{0.484 ± 0.059} & 0.219 ± 0.028 & 0.767 ± 0.006 & 0.691 ± 0.023 & 0.648 ± 0.511 & 0.993 ± 0.049 & 0.021 ± 0.060 & 1.044 ± 0.023 & -0.023 ± 0.033 \\
              & ald (Mode)    & 1.163 ± 0.340 &               & 0.750 ± 0.008 & 0.689 ± 0.023 &               &               &                &               &               \\
\midrule
              & ald (Mean)    & \textbf{0.363 ± 0.068} &               & 0.588 ± 0.014 & \textbf{0.585 ± 0.014} &               &               &                &               &               \\
LogNorm       & ald (Median)  & 0.533 ± 0.097 & 0.376 ± 0.013 & \textbf{0.589 ± 0.015} & 0.510 ± 0.023 & 0.256 ± 0.150 & 1.011 ± 0.028 & -0.004 ± 0.029 & 1.005 ± 0.021 & 0.006 ± 0.011 \\
              & ald (Mode)    & 1.733 ± 0.190 &               & 0.549 ± 0.043 & 0.496 ± 0.020 &               &               &                &               &               \\
\midrule
              & ald (Mean)    & 0.667 ± 0.139 &               & \textbf{0.919 ± 0.007} & \textbf{0.870 ± 0.029} &               &               &                &               &               \\
Norm heavy    & ald (Median)  & \textbf{0.454 ± 0.081} & 0.019 ± 0.001 & 0.916 ± 0.009 & 0.802 ± 0.008 & 0.256 ± 0.150 & 1.011 ± 0.028 & -0.004 ± 0.029 & 1.005 ± 0.021 & 0.006 ± 0.011 \\
              & ald (Mode)    & 0.627 ± 0.072 &               & 0.911 ± 0.012 & 0.802 ± 0.008 &               &               &                &               &               \\
\midrule
              & ald (Mean)    & \textbf{0.238 ± 0.036} &               & \textbf{0.894 ± 0.005} & \textbf{0.872 ± 0.004} &               &               &                &               &               \\
Norm med.     & ald (Median)  & 0.298 ± 0.036 & 0.047 ± 0.003 & 0.889 ± 0.006 & 0.868 ± 0.011 & 0.157 ± 0.044 & 0.997 ± 0.012 & 0.004 ± 0.014 & 1.058 ± 0.012 & -0.036 ± 0.011 \\
              & ald (Mode)    & 0.388 ± 0.047 &               & 0.884 ± 0.007 & 0.849 ± 0.011 &               &               &                &               &               \\
\midrule
              & ald (Mean)    & \textbf{0.236 ± 0.051} &               & \textbf{0.882 ± 0.004} & \textbf{0.874 ± 0.004} &               &               &                &               &               \\
Norm light    & ald (Median)  & 0.255 ± 0.016 & 0.090 ± 0.007 & 0.880 ± 0.003 & 0.853 ± 0.017 & 0.339 ± 0.076 & 0.998 ± 0.014 & 0.005 ± 0.021 & 1.087 ± 0.017 & -0.050 ± 0.011 \\
              & ald (Mode)    & 0.328 ± 0.029 &               & 0.876 ± 0.003 & 0.850 ± 0.017 &               &               &                &               &               \\
\midrule
              & ald (Mean)    & 0.404 ± 0.078 &               & \textbf{0.890 ± 0.005} & 0.847 ± 0.008 &               &               &                &               &               \\
Norm same     & ald (Median)  & \textbf{0.281 ± 0.022} & 0.066 ± 0.003 & 0.888 ± 0.006 & \textbf{0.886 ± 0.004} & 0.114 ± 0.036 & 1.004 ± 0.018 & 0.010 ± 0.022 & 1.007 ± 0.014 & 0.006 ± 0.012 \\
              & ald (Mode)    & 0.518 ± 0.065 &               & 0.881 ± 0.008 & 0.880 ± 0.004 &               &               &                &               &               \\
\midrule
              & ald (Mean)    & 0.385 ± 0.193 &               & 0.777 ± 0.012 & 0.727 ± 0.022 &               &               &                &               &               \\
LogNorm heavy & ald (Median)  & \textbf{0.244 ± 0.042} & 0.095 ± 0.006 & \textbf{0.779 ± 0.011} & \textbf{0.749 ± 0.011} & 0.043 ± 0.019 & 0.998 ± 0.014 & -0.002 ± 0.014 & 1.003 ± 0.014 & -0.005 ± 0.005 \\
              & ald (Mode)    & 0.898 ± 0.045 &               & 0.756 ± 0.029 & 0.724 ± 0.012 &               &               &                &               &               \\
\midrule
              & ald (Mean)    & \textbf{0.178 ± 0.046} &               & 0.747 ± 0.004 & 0.718 ± 0.007 &               &               &                &               &               \\
LogNorm med.  & ald (Median)  & 0.247 ± 0.024 & 0.174 ± 0.006 & \textbf{0.748 ± 0.004} & \textbf{0.749 ± 0.013} & 0.087 ± 0.052 & 1.002 ± 0.009 & -0.001 ± 0.012 & 1.008 ± 0.017 & -0.004 ± 0.010 \\
              & ald (Mode)    & 0.896 ± 0.082 &               & 0.723 ± 0.013 & 0.709 ± 0.012 &               &               &                &               &               \\
\midrule
              & ald (Mean)    & \textbf{0.184 ± 0.035} &               & \textbf{0.725 ± 0.007} & \textbf{0.713 ± 0.008} &               &               &                &               &               \\
LogNorm light & ald (Median)  & 0.221 ± 0.064 & 0.310 ± 0.011 & \textbf{0.725 ± 0.007} & 0.696 ± 0.020 & 0.185 ± 0.095 & 1.001 ± 0.014 & -0.001 ± 0.016 & 0.985 ± 0.015 & 0.008 ± 0.009 \\
              & ald (Mode)    & 0.921 ± 0.053 &               & 0.702 ± 0.014 & 0.697 ± 0.016 &               &               &                &               &               \\
\midrule
              & ald (Mean)    & \textbf{0.191 ± 0.044} &               & 0.739 ± 0.009 & 0.697 ± 0.008 &               &               &                &               &               \\
LogNorm same  & ald (Median)  & 0.259 ± 0.062 & 0.154 ± 0.006 & \textbf{0.740 ± 0.010} & \textbf{0.751 ± 0.014} & 0.076 ± 0.057 & 1.009 ± 0.011 & -0.005 ± 0.010 & 1.012 ± 0.011 & -0.001 ± 0.008 \\
              & ald (Mode)    & 0.943 ± 0.043 &               & 0.710 ± 0.007 & 0.715 ± 0.014 &               &               &                &               &               \\
\midrule
              & ald (Mean)    & 1.626 ± 0.194 &               & 0.637 ± 0.021 & \textbf{0.633 ± 0.031} &               &               &                &               &               \\
METABRIC      & ald (Median)  & 1.123 ± 0.088 & 0.245 ± 0.012 & \textbf{0.640 ± 0.018} & 0.588 ± 0.031 & 0.293 ± 0.125 & 0.993 ± 0.028 & -0.008 ± 0.024 & 1.001 ± 0.033 & -0.012 ± 0.016 \\
              & ald (Mode)    & \textbf{0.856 ± 0.039} &               & 0.605 ± 0.021 & 0.547 ± 0.018 &               &               &                &               &               \\
\midrule
              & ald (Mean)    & 2.196 ± 0.612 &               & \textbf{0.823 ± 0.016} & \textbf{0.824 ± 0.014} &               &               &                &               &               \\
WHAS          & ald (Median)  & 1.118 ± 0.152 & 0.134 ± 0.013 & 0.784 ± 0.043 & 0.765 ± 0.017 & 0.198 ± 0.094 & 0.981 ± 0.021 & 0.009 ± 0.023 & 0.972 ± 0.027 & 0.003 ± 0.016 \\
              & ald (Mode)    & \textbf{0.916 ± 0.101} &               & 0.802 ± 0.018 & 0.806 ± 0.022 &               &               &                &               &               \\
\midrule
              & ald (Mean)    & 1.121 ± 0.107 &               & 0.568 ± 0.015 & \textbf{0.572 ± 0.015} &               &               &                &               &               \\
SUPPORT       & ald (Median)  & 0.856 ± 0.062 & 0.362 ± 0.013 & \textbf{0.572 ± 0.015} & 0.561 ± 0.015 & 2.197 ± 0.667 & 0.900 ± 0.056 & 0.084 ± 0.046 & 1.084 ± 0.043 & -0.113 ± 0.023 \\
              & ald (Mode)    & \textbf{0.421 ± 0.051} &               & 0.532 ± 0.016 & 0.522 ± 0.044 &               &               &                &               &               \\
\midrule
              & ald (Mean)    & 1.713 ± 0.208 &               & 0.671 ± 0.014 & \textbf{0.665 ± 0.013} &               &               &                &               &               \\
GBSG          & ald (Median)  & 1.161 ± 0.094 & 0.278 ± 0.014 & \textbf{0.672 ± 0.010} & 0.590 ± 0.035 & 0.283 ± 0.106 & 0.977 ± 0.025 & 0.014 ± 0.034 & 1.000 ± 0.035 & -0.018 ± 0.016 \\
              & ald (Mode)    & \textbf{0.664 ± 0.072} &               & 0.657 ± 0.023 & 0.554 ± 0.062 &               &               &                &               &               \\
\midrule
              & ald (Mean)    & 3.002 ± 1.497 &               & 0.561 ± 0.037 & 0.547 ± 0.040 &               &               &                &               &               \\
TMBImmuno     & ald (Median)  & 1.085 ± 0.191 & 0.245 ± 0.015 & \textbf{0.562 ± 0.032} & \textbf{0.548 ± 0.030} & 0.835 ± 0.604 & 0.994 ± 0.021 & 0.004 ± 0.025 & 1.053 ± 0.045 & -0.038 ± 0.025 \\
              & ald (Mode)    & \textbf{0.609 ± 0.069} &               & 0.546 ± 0.024 & 0.531 ± 0.025 &               &               &                &               &               \\
\midrule
              & ald (Mean)    & 2.593 ± 0.289 &               & \textbf{0.617 ± 0.032} & \textbf{0.568 ± 0.036} &               &               &                &               &               \\
BreastMSK     & ald (Median)  & 1.116 ± 0.394 & 0.086 ± 0.008 & 0.457 ± 0.068 & 0.538 ± 0.083 & 0.066 ± 0.027 & 0.993 ± 0.020 & 0.003 ± 0.021 & 1.002 ± 0.019 & -0.007 ± 0.010 \\
              & ald (Mode)    & \textbf{0.686 ± 0.077} &               & 0.591 ± 0.071 & 0.515 ± 0.090 &               &               &                &               &               \\
\midrule
              & ald (Mean)    & 1.232 ± 0.325 &               & 0.778 ± 0.021 & 0.736 ± 0.030 &               &               &                &               &               \\
LGGGBM        & ald (Median)  & 0.846 ± 0.239 & 0.108 ± 0.011 & \textbf{0.785 ± 0.030} & \textbf{0.750 ± 0.043} & 0.450 ± 0.267 & 0.996 ± 0.038 & 0.008 ± 0.040 & 0.995 ± 0.047 & 0.003 ± 0.022 \\
              & ald (Mode)    & \textbf{0.497 ± 0.100} &               & 0.777 ± 0.023 & 0.739 ± 0.058 &               &               &                &               &               \\
\bottomrule
\end{tabular}
}
\end{table}

\begin{table}[ht]
\caption{The 50th, 75th and 95th percentiles of the CDF estimation for $t=0$, $F_{\text{ALD}}(0 \mid \mathbf{x})$, under the Asymmetric Laplace Distribution.}
\label{tab:ald_cdf}
\centering
\begin{tabular}{c c c c}
\toprule
\textbf{Dataset} & \textbf{50th Percentile} & \textbf{75th Percentile} & \textbf{95th Percentile} \\
\midrule
Norm linear       & 0.0001     & 0.0007     & 0.0018 \\
Norm non-linear   & 1.9878e-06 & 0.0001     & 0.0007 \\
Norm uniform      & 2.9879e-05 & 0.0028     & 0.0124 \\
Exponential       & 0.0194     & 0.0665     & 0.1204 \\
Weibull           & 0.0015     & 0.0032     & 0.0046 \\
LogNorm           & 0.0031     & 0.0109     & 0.0134 \\
Norm heavy        & 1.1804e-06 & 2.6128e-05 & 0.0007 \\
Norm med          & 4.2222e-06 & 3.5778e-05 & 0.0004 \\
Norm light        & 1.1978e-05 & 0.0001     & 0.0009 \\
Norm same         & 7.8051e-07 & 4.8624e-06 & 0.0001 \\
LogNorm heavy     & 0.0001     & 0.0014     & 0.0142 \\
LogNorm med       & 0.0001     & 0.0007     & 0.0082 \\
LogNorm light     & 0.0004     & 0.0024     & 0.0150 \\
LogNorm same      & 0.0004     & 0.0021     & 0.0123 \\
METABRIC          & 0.0068     & 0.0123     & 0.0292 \\
WHAS              & 0.0046     & 0.0151     & 0.0507 \\
SUPPORT           & 0.0957     & 0.1393     & 0.2035 \\
GBSG              & 0.0248     & 0.0394     & 0.0668 \\
TMBImmuno         & 0.0523     & 0.0681     & 0.0878 \\
BreastMSK         & 0.0006     & 0.0008     & 0.0130 \\
LGGGBM            & 0.0570     & 0.0842     & 0.1356 \\
\bottomrule
\end{tabular}
\end{table}


\end{document}